# Correcting Arabic Soft Spelling Mistakes using BiLSTM-based Machine Learning

A Preprint

**Gheith A. Abandah**
Computer Engineering Department
The University of Jordan
Amman, Jordan
abandah@ju.edu.jo

**Ashraf Suyyagh**
Computer Engineering Department
The University of Jordan
Amman, Jordan
a.suyyagh@ju.edu.jo

**Mohammed Z. Khedher**
Electrical Engineering Department
The University of Jordan
Amman, Jordan
khedher@ju.edu.jo

July 15, 2021

## Abstract

Soft spelling errors are a class of spelling mistakes that is widespread among native Arabic speakers and foreign learners alike. Some of these errors are typographical in nature. They occur due to orthographic variations of some Arabic letters and the complex rules that dictate their correct usage. Many people forgo these rules, and given the identical phonetic sounds, they often confuse such letters. In this paper, we propose a bidirectional long short-term memory network that corrects this class of errors. We develop, train, evaluate, and compare a set of BiLSTM networks. We approach the spelling correction problem at the character level. We handle Arabic texts from both classical and modern standard Arabic. We treat the problem as a one-to-one sequence transcription problem. Since the soft Arabic errors class encompasses omission and addition mistakes, to preserve the one-to-one sequence transcription, we propose a simple low-resource yet effective technique that maintains the one-to-one sequencing and avoids using a costly encoder-decoder architecture. We train the BiLSTM models to correct the spelling mistakes using transformed input and stochastic error injection approaches. We recommend a configuration that has two BiLSTM layers, uses the dropout regularization, and is trained using the latter training approach with error injection rate of 40%. The best model corrects 96.4% of the injected errors and achieves a low character error rate of 1.28% on a real test set of soft spelling mistakes.

## 1 Introduction

Arabic is one of the world's five major languages with over 290 million native speakers and a total of 422 million world speakers [1, 2]. Arabic is the fourth most common language on the Internet [3, 4] and the fastest growing language online [4]. A recent British Council report [5] ranks Arabic as the fourth needed language for the future based on economic and market factors, diplomatic and security priorities, mobility, tourism, and public interest. The Arabic language has 28 letters which we show alongside their Unicode encoding in Table 1. Of the 44 letters and accents listed, eight are letter variants: 0621–0626, 0629, and 0649. These variants are due to *hamzas* الهمزات (*al-hamzạt*) [1] placement and varying letter shapes at word endings. The three types of Arabic diacritics: *vowel*, *nunation* and *shadda* have the Unicode number range 064B–0652. They appear as accents above or below Arabic letters.

Western linguists distinguish between two forms of standard Arabic: Classical Arabic (CA) and Modern Standard Arabic (MSA). CA is the language of the Quran, ancient religious and liturgical texts, and old Arabic literature. MSA is the modern form that is syntactically, morphologically, and phonologically based on CA. MSA is the primary form of Arabic language used in education, business, media, news, and drafting laws and regulations. The contrast between MSA and CA "is mostly reflected in topic, vocabulary, and style rather than grammatical structure" [7]. Arabic is a diglossic language and Arabic linguists and speakers refer to both CA and MSA as *al-fusha* الفصحى to differentiate between the standard forms and the colloquial variants spoken throughout the Arab world. These dialects are not standardized and vary significantly socially and geographically. Furthermore, they are widely popular in use on social and messaging applications as well on the Internet [8]. For these reasons and many others, the U.S. state department categorizes Arabic among the exceptionally difficult languages to learn for native English speakers [9]. Consequently, it is vital to develop and modernize automated tools that aid native speakers and learners alike in communicating in Arabic using correct grammar and spelling, especially in online communication. To this end, our previous work [10, 11, 12] investigated machine learning and hybrid approaches for Arabic text diacritization with recurrent neural networks (RNN). This novel work investigates Arabic text correction using RNN in contrast to traditional techniques.

Research efforts have normally focused on a subset of the typical spelling and writing mistakes encountered in Arabic texts; usually by addressing one or two at a time. These errors have traditionally been categorized into different types [13, 14]: discourse structure errors, pragmatic-level errors, morpho-syntactic, lexical, context (semantic or real-word) errors, and soft errors.

### 1.1 Discourse Structure and Pragmatic-level Errors

Discourse structure errors are analyzed at the sentence and paragraph levels. The word flow is free of grammatical and misspelling errors but the overall communication suffers from incoherence and illogicality. In contrast, pragmatic errors are due to misunderstanding of the implied word meaning within the context. It plays a crucial part in language translation.

[1] The transliteration used in this paper is Romanization using Glosbe transliteration tools [6].

Table 1: Unicode Arabic code block

| x | U+062x | Name | U+063x | Name | U+064x | Name | U+065x | Name |
|---|---|---|---|---|---|---|---|---|
| **0** | | | ذ | *thal* | | | ـِ | *kasra* |
| **1** | ء | *hamza* | ر | *reh* | ف | *feh* | ـّ | *shadda* |
| **2** | آ | *alef/madda above* | ز | *zain* | ق | *qaf* | ـْ | *sukun* |
| **3** | أ | *alef/hamza above* | س | *seen* | ك | *kaf* | | |
| **4** | ؤ | *waw/hamza above* | ش | *sheen* | ل | *lam* | | |
| **5** | إ | *alef/hamza below* | ص | *sad* | م | *meem* | | |
| **6** | ئ | *yeh/hamza above* | ض | *dad* | ن | *noon* | | |
| **7** | ا | *alef* | ط | *tah* | ه | *heh* | | |
| **8** | ب | *beh* | ظ | *zah* | و | *waw* | | |
| **9** | ة | *teh marbuta* | ع | *ain* | ى | *alef maksura* | | |
| **A** | ت | *teh* | غ | *ghain* | ي | *yeh* | | |
| **B** | ث | *theh* | | | ـً | *fathatan* | | |
| **C** | ج | *jeem* | | | ـٌ | *dammatan* | | |
| **D** | ح | *hah* | | | ـٍ | *kasratan* | | |
| **E** | خ | *khah* | | | ـَ | *fatha* | | |
| **F** | د | *dal* | | | ـُ | *famma* | | |

### 1.2 Morpho-syntactic Errors

Morpho-syntactic errors occur when there is a disagreement in gender, number, wrong verb tense, or wrong word ordering. Given that Arabic is a Semitic language, it uses dual nouns and verbs for both the feminine and masculine aside from the respective plural forms for both genders. As such, a common morpho-syntactic error is verb and number disagreement. For example, when writing "The two men are eating lunch", one might erroneously write الرجلان يتناولوا الغداء (*ạl-rajulạn yatanạwalw ạl-gẖadạʾ*) whereas the correct way is to write الرجلان يتناولان الغداء (*ạlrajulạn yatanạwalạn ạlgẖadạʾ*). The dual form of 'man' must be followed by the dual form of the verb 'eat' يتناولان (*yatanạwalạn*). In the previous example, the syntactic error is to use the plural form of the verb 'eat' يتناولوا (*yatanạwalw*).

### 1.3 Lexical and Semantic (Context) Errors

Lexical errors are the spelling mistakes that result in non-existent words in the lexicon. Context or semantic errors include spelling mistakes which render another valid and existing word which is out of the context of the sentence. One source of lexical and semantic errors is mistranscription between phonetics (speech) and the written typographic version of the word. The word 'conscience' might be mistakenly written as دمير (*damyr*) instead of ضمير (*ḍamyr*) where the former does not exist in the Arabic lexicon. Furthermore, the letter *lam* in the definitive prefix *al-* comes in two distinctive phonetic forms called the solar and lunar *lam* لام شمسية ولام قمرية (*lạm sẖamsyạ̃ẗ wlạm qamaryạ̃ẗ*). The solar *lam* is dropped in pronunciation leading to phonetic transcription errors. In writing the word 'the sky' السماء (*ạl-sãmạʾ*) that is pronounced as (*ạsãmạʾ*), it is natural to misspell it as it is pronounced اسماء (*ạsãmạʾ*).

Another source of lexical and semantic errors is typographical mistakes due to keyboard mistyping, or erroneous optical character recognition (OCR). Diacritics (vowel marks) might confuse the OCR system into recognizing one letter for another. For example, the last word in the sentence 'He is injured' إنه جريحٌ (*ạinãhu jaryḥu*$^{n}$) can be mistakenly written as إنه جريخ (*ạinãhu jarykẖ*). The *dammatan* diacritic *(u*$^{n}$*)* above the *hah* letter rendered it to be mistaken as letter *khah.*

### 1.4 Soft Errors

Soft spelling mistakes are a special type of lexical and semantic errors that are due to the orthographic variations of some Arabic letters. For example, at the beginning of a word, the Arabic letter *alef* - comes in bare *alef* and and *hamza* forms: *alef with madda above*, *alef with hamza above* and *alef with hamza below.* The former bare *alef* shape is called همزة وصل (*hamzaẗ waṣl*) while the *alef* with *hamza* above or below is called همزة قطع (*hamzaẗ qaṭ*$^{c}$). Replacing one form of *alef* with another is a major case of soft-spelling mistakes [15].

Despite having standard rules for the shape and placement of the *hamza* inside the word, or as known in Arabic الهمزة المتوسطة (*ạl-hamzaẗ ạl-mutawsiṭaẗ*); these rules are quite complex for natives and learners alike. We can write the middle *hamza* above an *alef* as in the noun 'head' رأس (*rȃs*), or above the letter *waw* as in the noun 'vision' رؤية (*rûyaẗ*), or on a mark نبرة (*nabirẗ*) as in the noun 'well' بئر (*bŷr*), or standalone *hamza* form as in the noun 'reading' قراءة (*qirạʾaẗ*). We can determine the correct spelling by examining the diacritics of the *hamza* itself and those of the preceding letter. Some exceptions do apply. Similar Arabic spelling rules dictate how we write the *hamza* at the end of a word, which is called الهمزة المتطرفة (*ạl-hamzaẗ ạl-mutaṭarifaẗ*). We can write the *hamza* above an *alef* as in the noun 'refuge' ملجأ (*maljȃ*). We can place it above a *waw* as in the verb 'to dare' يجرؤ (*yajrû*) or over the *alef maksura* as in the noun 'ports' الموانئ (*ạl-mawạnŷ).* Finally, it can rest alone as in the noun 'warmth' دفء (*difʾ*). Clearly, the shape that the *hamza* takes based on its possible placement within a word could be quite confusing.

There are other notorious examples in the soft misspellings category that are not necessarily related to the *hamza* . These include inserting or omitting the *alef* following a *waw* letter at the end of a word. A common example for the insertion error is adding an *alef* at the end of present tense verbs that end with a *waw*; such as writing the verb 'to kneel' as يجثوا (*yajtẖwạ*) instead of يجثو (*yajtẖw*). We frequently encounter the omission error while spelling conjugated imperative plural verbs. For example, in the imperative sentence 'Don't look' addressing a group of people, the verb can by incorrectly spelled as لا تنظرو (*lạ tanẓurw*) instead of لا تنظروا (*lạ tanẓurwạ*).

Table 2: The frequency of soft spelling errors analyzed in Al-Ameri study [16]

| No. | **Spelling Error Type** | % | Wrong Spelling | Correct Spelling |
|---|---|---|---|---|
| 1 | Writing *middle hamza* on an *alef* | 73% | قرأة (*qirảẗ*) | قراءة (*qiraᵓẗ*) |
| 2 | Writing *alef maksura* instead of *alef* | 71% | عفى (*ᶜafy*) | عفا (*ᶜafạ*) |
| 3 | Writing *alef* instead of *alef maksura* | 70% | بكا (*bakạ*) | بكى (*baky*) |
| 4 | Omitting *alef* following a *waw* at the end of a verb | 67% | لا تنظرو (*lạ tanẓurw*) | لا تنظروا (*lạ tanẓurwạ*) |
| 5 | Writing *teh* instead of *teh marbuta* | 67% | كثرت (*kat͟hrat*) | كثرة (*kat͟hrẗ*) |
| 7 | Writing *middle hamza* on *waw* | 64% | المروؤة (*ạl-murww̉aẗ*) | المروءة (*ạl-murwᵓaẗ*) |
| 8 | Writing *hamza* at the end of the word on *alef* | 51% | أشيأ (*ảs͟hyả*) | أشياء (*ảs͟hyạᵓ*) |
| 9 | Writing a standalone *hamza* at the end of the word | 47% | خطء (*k͟hạṭᵓ*) | خطأ (*k͟hạṭả*) |
| 10 | Writing *hamza* at the end of the word on *yeh* | 47% | شئ (*s͟hạỷ*) | شيء (*s͟hạyᵓ*) |
| 11 | Dropping *lam* before a "solar letter" | 38% | اسماء (*ạsmạᵓ*) | السّماء (*ạl-sãmạᵓ*) |
| 12 | Writing *teh marbuta* instead of *teh* | 37% | ساعاة (*sạᶜạẗ*) | ساعات (*sạᶜạt*) |
| 13 | Writing *middle hamza* on *yeh* | 30% | يتفائل (*yatafạᵓal*) | يتفاءل (*yatafạỷl*) |
| 14 | Writing *hamzaẗ qaṭᶜ* instead of *hamzaẗ waṣl* | 28% | إبن (*ạbn*) | ابن (*ạbn*) |
| 15 | Inserting *alef* after *waw* at the end of a word | 25% | يجثوا (*yajt͟hwạ*) | يجثو (*yajt͟hw*) |
| 16 | Confusing *teh marbuta* and *heh* at the end of a word | * | مكتبه(*makatabh*) | مكتبة (*maktabaẗ*) |

* Common mistake yet not reported in this study

Furthermore, common soft misspellings include confusing the *teh marbuta* at the end of the word for a *teh*. One example is writing the noun 'watches' as ساعاة (*sạᶜạẗ*) instead of ساعات (*sạᶜạt*). The *teh marbuta* is also often confused for the letter *heh* as in the word for library مكتبه/مكتبة (maktabaẗ/maktabah). Finally, due to the similar shapes between the letters for *alef maksura/yeh* written at the end of Arabic words, we can see the letters mistakenly interchanged as in the adjective 'interactive' تفاعلى/تفاعلي (*tafạᶜuly /tafạᶜuly*). The correct shape of the *alef* at the end of a word is dictated by grammatical rules based on the Arabic word root system. Many people forgo the rules, and given the identical phonetic sounds, the *alef maksura* and the *alef* are often written one for the other. To illustrate, the correct spelling for the singular masculine past tense of the verb 'to cry' is بكى (*baky*) where it is often misspelled as بكا (*bakạ*). In contrast, the singular masculine past tense of the verb 'to forgive' is عفا (*ᶜafạ*) and it is misspelled as عفى (*ᶜafy*).

Al-Ameri [16] analyzed the frequency of Arabic spelling mistakes in a sample of Teacher Education Institutes attendees. We notice that the majority of reported errors relate to typographical errors due to phonetic mistranscription (*i.e.*, mistaking a diacritic for a letter or vice versa, phonetically close letters); errors due to *hamza*; and finally confusing the end *alef* or *teh* for other orthographic forms or letters. Table 2 summarizes the most common soft-spelling errors encountered in Al-Ameri's study.

It is worth noting that the lack of a common and sufficiently large enough benchmark dataset for Arabic spelling errors has hindered the continuous progress in the field of Arabic spelling errors detection and correction. This holds for both classical and modern standard Arabic sets. Many of the existing sets suffer from the lack of proper and comprehensive annotation, variety, and consistency. For example, within the same set, one can find texts that are fully diacritized while others have minimal or lack diacritization altogether. The lack of annotation makes the tasks of mapping the dataset texts to grammatically sound and misspellings-free texts difficult. Many times, researchers spend cumbersome manual effort in preparing sets for study. Other times, it is easier to introduce artificial mistakes from known correct texts. While this could work most of the time; the injection of artificial errors might not necessarily correspond to the real errors made by language speakers and learners alike.

### 1.5 Approach and Contribution

In this paper, we propose using a tuned bidirectional long short-term memory (BiLSTM) recurrent neural network to detect and correct spelling mistakes written in either classical or modern standard Arabic. We target a subset of the most commonly encountered spelling mistakes in the Arabic language [16]. Mainly, errors in the soft misspelling category pertaining to *ạl-hamzạt* (الهمزات, the different shapes of *alef*), and the common errors in shaping *teh* at the end of the word. We propose tackling the problem at the character-level and we propose letter conversion scheme that allows one-to-one training of the input sequences against the target sequences.

We propose and evaluate two approaches to train models to correct these mistakes. In the *transformed input* approach, the network is trained to predict correct spelling from transformed unified input. Whereas the *stochastic error injection* trains the network to correct randomly injected spelling mistakes. We recommend best configuration and approach based on evaluation on two training datasets and a sample of real mistakes.

We organize the rest of the paper as follows: In Section 2, we survey the state-of-the-art techniques in detecting and correcting spelling mistakes in European, Indo-Iranian and the Arabic languages. In Section 3, we review the basic concepts of recurrent neural networks, long short-term memory, and the sequence transcription problem. Section 4 details the experimental setup used in this work, the datasets, the used training approaches, and the performance evaluation metrics. Section 5 presents and discusses the results of our experiments. We provide a summary and conclude the paper in Section 6.

## 2 Related Work

The natural language processing community has an ongoing interest in spell checking and correction. Traditionally, post-OCR spell check and correction has been a driving force behind research and application. Yet, in the past decade, the proliferation of social media and the high reliance on instant messaging demand more efficient and accurate spell checkers and on-the-fly accurate correction. The accuracy level of spelling and grammatical mistakes correction varies in maturity between different languages. It is worth noting that research experiments have been usually evaluated based on artificially-created or proprietary corpora and less so on a corpus of authentic misspellings [17]. Moreover, generic spelling checkers (GSCs), such as the ones packaged in popular text editors like Microsoft Word, are designed for native writers and as such fail to detect and correct mistakes commonly introduced by second language learners [18]. In an ever-interconnected world where hundreds of millions of people are bilingual and multilingual, designing accurate spell-checkers is more challenging. In this section, we review some of the most recent works in spelling correction for three world language groups.

### 2.1 Spelling Correction for European Languages

Whereas spelling correction for the English language is well-established compared to other languages, most of this research was evaluated on proprietary corpora of native English texts or well-formed texts with artificially injected errors. Yet, spelling correction of non-native speakers is far more challenging as they feature multi-character edits compared to a single-character edit produced by native speakers [19]. To this end, the authors in [17] developed a minimally supervised model based on contextual and non-contextual features. These features include orthographic similarity, phonetic similarity, word frequency, n-grams and word embeddings among others. Notably, they present and use a corpus of real-world learner essays from the TOEFL exam, and further test their model on out-of-domain medical notes. They report an accuracy level of 88.12% on the TOEFL set, and 87.63% on the medical set. Li *et al.* [20] propose a nested RNN model for English word spelling error correction. They generate pseudo data based on phonetic similarity to train the network. Their proposed system has a precision of 71.77%, a recall rate of 61.26%, and an $F_{0.05}$ score of 69.39%.

D'hondt *et al.* [21] employ many-to-many character sequence learning network using LSTM for French text correction. Their model stacks two LSTM layers: an encoder layer that reads the sequence of characters, and a decoder layer that generates the output. They further use a drop-out layer to enhance performance. They train and evaluate their system on a dataset of OCRed French medical notes using two models that introduce noise and confusion into the text. They report an accuracy rate of 73% for the former and 71% for the latter model. The same authors extended their work by using BiLSTM [22], and used various corpora based on structured English, structured French, and free-text French with artificially corrupted strings. They report accuracy rates no less than 85% for the structured English and French, and 60% for the free-text French. They show that for an original text with a character error rate (CER) of 34.3%, the BiLSTM system reduces the CER to 7.1%.

### 2.2 Spelling Correction for Indo-Iranian and Asian Languages

Dastgheib *et al.* [23] introduced Perspell; a semantic-based spelling correction system for the Persian language which is based on an n-gram model. Perspell handles both real-word and non-word errors. The authors' experiments show that for non-word errors, the precision, recall, and $F_1$ score are 87.7%, 88.9%, and 88.3%, respectively; while for real-word errors, the authors report a precision of 92.4%, a recall rate of 93%, and an $F_1$ score of 92.6%.

More recently, Yazdani *et al.* [24] use dictionary-based methods to detect word misspells. They rely on a generic Persian dictionary and a specialized medical dictionary as their system is oriented towards health care applications, specifically ultrasound reports. They employ an n-gram model to dictate suggestions based on orthographic and edit distances. They test their system on actual ultrasound free-text reports and achieve a detection performance of up to 90.29% and a correction accuracy of 88.56%.

Salavati *et al.* [25] introduce Rênûs, a spell checker for the Sorani dialect of the Kurdish language. The error detection phase in Rênûs is based on an n-gram frequency model, and the error correction phase is based on the edit distance, as a measure of similarity and frequency. The authors carried out experiments to investigate error correction once with the use of lexicon and another without. They report a correction accuracy of 96.4% for the former and 87% for the latter case.

The work in [26] introduces the SCMIL system which stands for sequence-to-sequence text correction model for Indic languages. SCMIL uses an attention model with a bidirectional RNN encoder and attention decoder. The decoder is trained end-to-end and it has a character-based representation on both encoder and decoder sides. They have synthesized a dataset from the Hindi and Telugu languages with data lists comprised of a maximum of five words. They subsequently introduced errors which include insertion, deletion, substitution, and word fusion. The authors show that SCMIL has an accuracy rate of 85.4% for the Hindi language and 89.3% for the Telugu language.

Zhang *et al.* proposed a system for Chinese spelling error detection which consists of a network for error detection and a network for error correction based on BERT. The two networks are connected to each other through a technique they called soft-masking. For a training set of five million examples, the authors' error detection system achieved an accuracy of 80.8%, a precision of 65.5%, a recall of 64% and an $F_1$ score of 64.8%. The error correction system; however, achieved an accuracy of 77.6%, a precision of 55.8%, a recall of 54.5% and an $F_1$ score of 55.2%.

### 2.3 Spelling Correction for the Arabic Language

Most recent works for Arabic spelling detection and correction still use traditional techniques in the field of natural language processing (NLP). For example, Mars [27] introduced a spell checker which targets both lexical and semantic spelling mistakes. He uses a sequential combination of approaches including lexicon-based, rule-based, and statistical-based methods. He achieves an $F_1$ score of 67%. Al-Shneifi *et al.* [28] developed a cascade system called Arib that detects and corrects a range of spelling errors. Errors that are discovered by Arib include: edit, add, split, merge, punctuation, phonological, and other observed common mistakes. They employed two core models: a probabilistic model based on Bayes probability theory and a Levenshtein distance-based model. They further add three extra models; two of which are based on 3$^{rd}$ party error detection tools: MADAMIRA and Ghaltawi, and the third additional module is a rule-based correcter derived from analyzing samples of the QALB database. Overall, Arib has an $F_1$ score of 57.8%, and precision and recall rates of 66.6% and 51.1%, respectively. Mubarak and Darwish [29] also used a cascaded approach for word-level errors, followed by punctuation correction. For word-level correction, the authors used a statistical character-level transformation model and a language model to handle letter insertions, deletions, and substitutions and word merges. The author subsequently use a case-specific system aided by a language model to handle specific error types such as dialectal word substitutions and word splits. For punctuation recovery, the authors employ a simple statistical word-based system and a conditional random fields sequence labeler (CRF). For different experiments, the authors were able to achieve a precision rate up to 71.7%, a recall rate up to 60.32%, and an F-measure up to 63.43%.

Bouamor [30] introduced another hybrid system that is based on a morphology-based corrector; rule-based linguistic techniques, language modeling, statistical machine translation (SMT), as well as an error-tolerant finite-state automata method. They target common error types which include split, delete, edit, merge, move and add errors based on the 2014 QALB set. They report an $F_1$ score of 68.4%.

Noaman *et al.* [31] developed a hybrid system based on the concept of confusion matrix and the noisy channel spelling correction model. They automatically detect and correct Arabic spelling errors of the edit and split types based on the QALB dataset. They report a word error correction rate up to 89.7%. Zahui *et al.* [14] introduced Al- Mossahih tool that detects and corrects one-letter typographical and phonetic transcription errors. Their detection phase is dictionary-based where a collection of around two million words are sorted in alphabetical order. The correction module encompasses four techniques: one is based on a correspondence table between pairs of commonly confused characters, the second is permutation-based where all possible words from the word letters are generated, the third is a neighborhood module which considers letters whose keys are nearby on the keyboard, and finally a language model that deals with word locations within a sentence. Al-Mossasih tool has a word error detection rate of 74.75% and a correction rate of 80.2%.

Semantic errors have been addressed by few recent works. A major approach is based on confusion matrices, which despite being powerful, they limit the number of errors that can be detected and corrected. Al-Jefri and Mahmoud [32] compiled a corpus of 7.4 million words from the set of words most confused by non-native Arabic speakers and from the set of mis-recognized words by Arabic OCR systems. The authors compiled these words into 28 confusion sets with assigned probabilities. Errors detection only targets words listed in the confusion matrices and error correction is based on picking the word with highest probability using the computed n-gram model. They report an average accuracy of 95.4%.

For non-confusion set-based approaches, Zribi and Ahmed [33] detected semantic errors through the use of four combined statistical and linguistic methods. They have introduced semantic errors on a set of sentences collected from economic articles from the Egyptian Al-Ahram newspaper. The semantic errors were all a single edit away from the correct word. The reported detection performance was 90% and 83% for precision and recall, respectively. Rokaya [34] introduced a small variation into the previous method by using the power link method instead of traditional frequency to detect and correct semantic errors coupled with confusion sets as a hybrid approach. They only report results for the detection stage where their system achieves 94.35% and 85.57% for precision and recall, respectively.

More recently, Watson *et al.* [35] utilized sequence-to-sequence models and character and word embeddings for Arabic Text Normalization. Azmi *et al.* [13] combine the language model with machine learning in the detection stage. For the correction step, they only use a language model. They have used word n-grams as features which are subsequently fed into a support vector machine classifier (SVM) to detect and mark words with semantic errors. For the correction step, they generate candidate words which are one-edit distance away from the erroneous word. The candidates are ranked and sorted based on the n-gram language model and then they select the best suggestion accordingly. Their system has an $F_1$ score of 90.7%, and an precision and recall rates of 83.5% and 99.2%, respectively. Alkhatib *et al.* [36] recently used an LSTM model to detect and correct spelling and grammatical mistakes at the *word-level*. Their model uses word-embeddings and a polynomial classifier. They report an $F_1$ score of 93.89%, and for morpho-syntactic mistakes pertaining to word form, noun number, verb form, and verb tense, they report a precision of 95.6% and a recall rate of 94.88%. Solyman *et al.* [37] employed CNNs for the automatic correction of Arabic grammar. After fine-tuning their different developed and tested models, the authors achieved a precision of 80.23%, a recall rate of 63.59%, and an $F_1$ score of 70.91. Kuznetsov and Urdiales [38] proposed a method of performing spelling correction on short input strings, such as search queries or individual words using denoising auto-encoder transformer model to recover the original query. They used datasets for four langauges and achieved an accuracy of 83.33% (Arabic), 91.83% (Russian), 93.97% (Greek), and 94.48% (Setswana).

## 3 Machine Learning and Sequence Transcription

A general definition of sequence transcription is the process of transforming an input sequence into a corresponding output sequence. Within the context of machine learning spelling detection and correction, the input sequence is the set of letters forming the text that may have spelling errors. The corresponding output sequence is the text on which the machine learning algorithm attempted corrections. Sequence transcription is quite common in similar problems in the fields of language translation, voice recognition and diacritizing Arabic texts [12]. In all these applications, we need to infer relationships and provide outputs depending on past input data. Therefore, the need to preserve correlations between data points in the sequence is necessary. Recurrent neural networks (RNN) provide the capability to learn from data sequences and consequently infer relevant output data. In this section, and for the sake of completeness, we briefly review RNN and a special RNN network cell called long short-term memory (LSTM) that we adopted in this work. We will further provide details of how we handle and process the input sequence prior to its use in the LSTM network.

### 3.1 Basic RNN

In general, recurrent neural networks (RNN) maintain hidden states that are functions of previous input. This enables such networks to infer outputs from past sequences making them quite suitable for applications where we require sequence transcription. A standard RNN cell can be described by two equations that relate the input sequence $x_t \in (x_1, x_2, \ldots x_T)$ to the hidden states $h_t \in (h_1, h_2, \ldots h_T)$ and output sequence

$y_t \in (y_1, y_2, \ldots y_T)$. Eqs. 1 and 2 describe the standard RNN model:

$$h_t = \sigma(W_h x_t + U_h h_{t-1} + b_h) \tag{1}$$

$$y_t = W_y h_t + b_y \tag{2}$$

where $W$, $U$, and $b$ denote the weight and bias matrices. We can clearly observe that the current hidden state $h_t$ depends on both the current input $x_t$ and the previous hidden state $h_{t-1}$. The model uses a sigmoid function $\sigma$ to limit the output within a prefixed range. Hyberbolic tangents *tanh*, for example, limit the output to between $-1$ and 1, whilst logistic sigmoid limits the output to between 0 and 1. Both functions are often used in RNN among others. In this paper, we strictly use the $\sigma$ symbol to denote a logistic sigmoid whilst we express the hyperbolic tangent as *tanh*. As stated earlier, the hidden state equips the network model with the ability to learn from input sequences, remember past data and infer outputs. However, if the input sequence is long, and the output data to be inferred depends on a much older (past) input data sequence (*i.e.*, the gap or the distance in sequence between the inferred output and relevant input data is large), RNN in its standard form might not be able to deliver an accurate desired output sequence. To this end, researchers developed the refined RNN cell which they called long short-term memory (LSTM) [39]. LSTM can handle much longer sequences as well as avoid an inherent problem in standard RNN: the vanishing gradient problem. The vanishing gradient problem halts the update of network weights when the gradient is too small; thus stopping the learning stage quite early resulting in a poor network model.

### 3.2 Long Short-Term Memory RNN (LSTM-RNN)

The internal architecture of an LSTM cell vastly improves on the basic RNN cell by incorporating a *set* of gates which govern the operation of each individual cell. These gates equip the cell with the capacity to work with short or long-term contexts. LSTM cells are relatively insensitive to long gaps or large distance between the output to be inferred and relevant old data in the sequence. Figure 1 illustrates the internal architecture of an LSTM cell. An LSTM cell has an *input* gate **I**, a *forget* gate **F**, an *output* gate **O**, and a *cell activation* unit **C**. We provide the governing equations of these gates in Eqs. 3 through 7. For each gate type, we use the small letter notation to denote the output of the gate at time $t$:

$$i_t = \sigma(W_i x_t + U_i h_{t-1} + V_i c_{t-1} + b_i) \tag{3}$$

$$f_t = \sigma(W_f x_t + U_f h_{t-1} + V_f c_{t-1} + b_f) \tag{4}$$

$$c_t = f_t \circ c_{t-1} + i_t \circ tanh(W_c x_t + U_c h_{t-1} + b_c) \tag{5}$$

$$o_t = \sigma(W_o x_t + U_o h_{t-1} + V_o c_t + b_o) \tag{6}$$

$$h_t = o_t \circ tanh(c_t) \tag{7}$$

where $W$, $U$, $V$, and $b$ denote the weight and bias matrices and the initial values for $c_0 = h_0 = 0$. The $\circ$ operator denotes the Hadamard (element-wise) product. Similar to the basic RNN cell, an update to a short-term state $h_t$ or long-term state $c_t$ depends on the current input $x_t$, and previous short-term and long-term memory states $h_{t-1}$ and $c_{t-1}$.

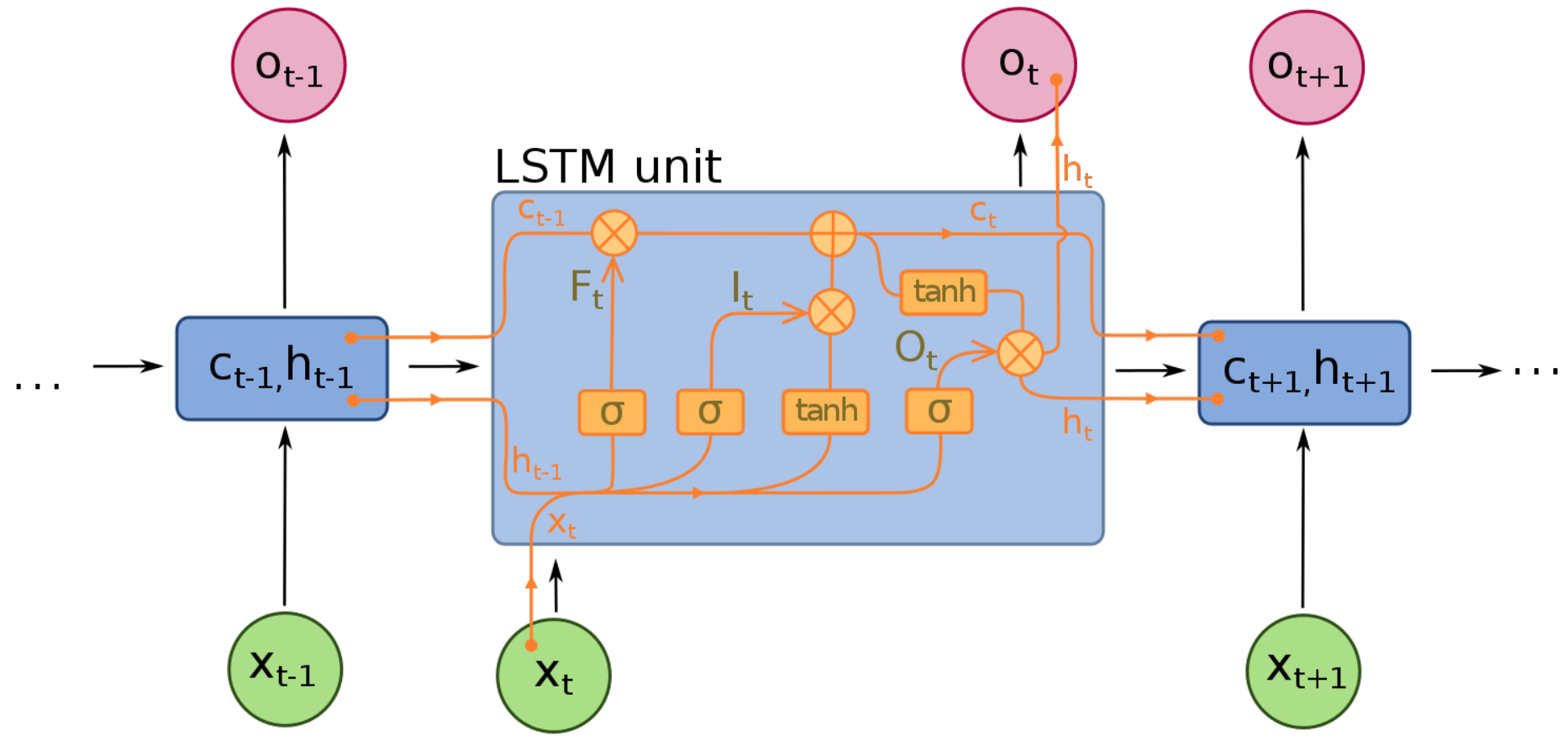

Figure 1: Basic structure of an LSTM cell [40]

LSTM is likely to quickly over-fit the training data; thus rendering them less powerful in predicting correct outputs. Dropout is a computationally cheap way that reduces over-fitting. It further improves generalization error and model performance in all deep neural networks. Dropout simply implies probabilistically excluding nodes from activation and weight updates while training a network.

The bidirectional LSTM is a version of the LSTM architecture that exploits future contexts as well as past contexts. That is, in sequence transcription problems, it could be beneficial to have access to subsequent (future) sequences to infer correct current outputs. In BiLSTM, we couple the conventional unidirectional LSTM network that works with past sequences with another unidirectional network that works with future sequences. We train the former in a forwards fashion whilst the latter in a backwards fashion. The final output is a concatenation of both the forward and backward LSTM networks.

### 3.3 Sequence Processing and Data Encoding

We can classify RNN based on the relationship between the input and output sequence lengths into four types:

- *One-to-one* networks where the length of the input sequence matches the length of the output sequence.
- *Many-to-one* networks where input sequences are transcribed into one final output, for example classification problems.
- *One-to-many* networks where one input vector is used to produce an output sequence.
- *General many-to-many* networks where the number of items in the input sequence differs than that in the output sequence. We use encoder-decoder architectures to tackle these transcription problems by using an intermediary fixed-length vector. The encoder maps a variable-length source sequence to the intermediary fixed-length vector, and the decoder maps the vector representation to a variable-length target sequence.

We can model the Arabic spelling correction problem as a one-to-one problem. Intuitively, this should be straightforward as most soft Arabic spelling errors result from confusing shapes of الهمزات (ạl-hamzạt) or the similar sounding characters at the end of the word. Simple one-character replacement fixes the misspelled word. However, some of these soft mistakes fall under the addition/omission type. For example, the spelling error in cases 4 and 15 in Table 2 add or omit the *alef* at the end of the word after a *waw*. Similarly, case one corrects writing *middle hamza* on an *alef* (one letter) by writing it properly following an *alef* (two letters). These particular cases result in differing-length sequences. Ordinarily, the encoder/decoder RNN architecture handles this well, yet with extra cost and overhead.

To mitigate and simplify our approach, we propose a simple yet effective technique that maintains one-to-one sequencing by processing the input sequence stream prior to applying it to the neural network, and then post-processing the output sequence to restore readable Arabic form. We convert some letters and two-letter combinations to intermediate arbitrary codes of English letters, as specified in Table 3. The conversion involves letters that are prone to the spelling mistakes under study: the combination *alef-hamza* (اء) and the letters at word ends (ـوا, ـو, ـت and ـه). We emphasize that this conversion is positional in nature; that is; we convert the *waw-alef* combination (ـوا) to an 'A' only when it appears at word endings where the associated error frequently occurs. Should this combination appear in the middle, no conversion is performed. However, we convert the letter combination *alef-hamza* to 'J' wherever it occurs in the word. This combination is susceptible to soft mistakes both in the middle and at the end of the word.

Table 3: Letters converted to intermediate codes

| Letter(s) | Intermediate Code |
|---|---|
| اء | J |
| ـوا | A |
| ـو | O |
| ـت | T |
| ـه | H |

In our machine training approach, we use this converted sequence as the target sequence and a copy of it as the input sequence after injecting some artificial spelling mistakes (refer to Section 4.2). These sequences are stored using the Unicode UTF-8 encoding. However, when presented to the neural network, they are put in 3D ($B \times T \times C$) tensors (dense matrices). Each tensor holds a batch of $B$ sequences of a maximum length $T = 400$ characters. Note that we wrap sequences longer than 400 characters to improve training performance. The third dimension encodes each of the $C$ distinct characters using one-hot encoding.

# 4 Experimental Setup

In this work, we use an experimental setup similar to the one used in our past research [41]. We list the specifications of the experimental platform in Table 4.

Table 4: Experimental Platform Specifications

| **Aspect** | **Specification** |
|---|---|
| CPU | Intel Core i7-9700KF @ 3.6 GHz, 8 cores, 12 MB cache |
| GPU | Nvidia GeForce RTX 2080 @ 2.1 GHz, 2944 CUDA cores, 8 GB memory |
| Memory | 32 GB DDR4-SDRAM @ 2666 MHz |
| OS | Ubuntu 20.04 LTS, 64-bit |
| Libraries | Python 3.8.2, TensorFlow 2.2.0, Keras 2.3.0-tf |

## 4.1 Machine Learning Model Configurations

We develop our machine learning models using Python deep learning libraries ensuring we use the latest versions of the algorithms. Specifically, we use Keras with TensorFlow at the backend. Given that we tackle the problem of Arabic spelling correction at the character level, we adopt BiLSTM RNN. These networks can handle longer sequences which can be beneficial for correcting misspelled words based on past and future context. We develop and compare three models. The baseline model has only two BiLSTM layers. We add an input masking layer before the two hidden BiLSTM layers, and connect their output to a fully-connected (dense) output layer, as shown in Fig. 2.

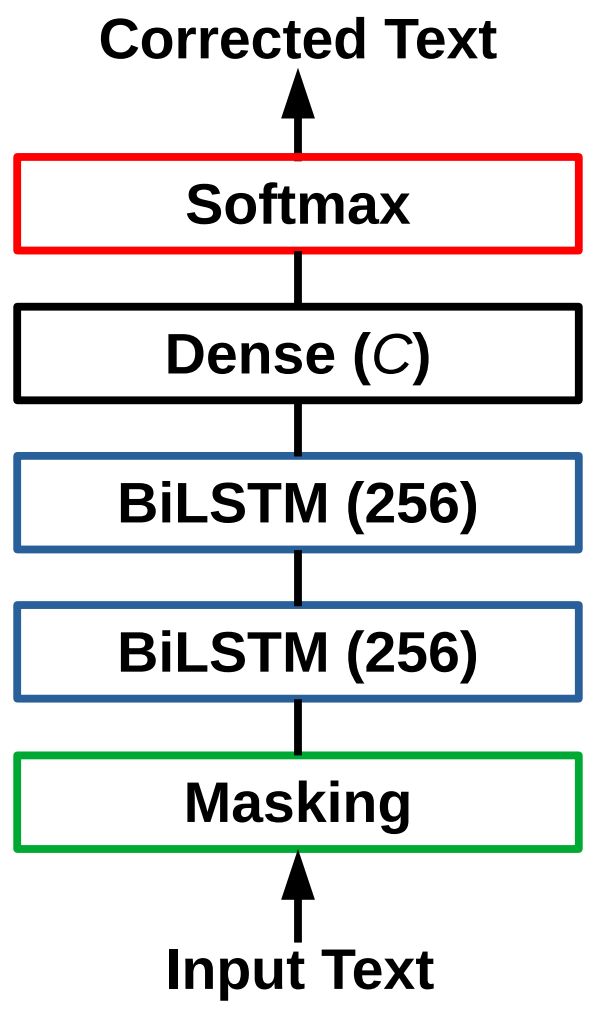

Figure 2: The Base 2-Layer Network Model

The second model maintains the same settings of the previous one but further employs the *dropout* method to reduce or avoid the over-fitting problem. The third and final model has four BiLSTM hidden layers instead of two and also uses *dropout*. We use the same configuration settings for the three models. Each

Listing 1: A 2-Layer BiLSTM Model with Dropout

```
model = Sequential()
model.add(Masking(mask_value = 0,input_shape=(seq_len, num_inp_tokens)))
model.add(Bidirectional(LSTM(256, return_sequences=True,dropout=0.1, recurrent_dropout
    =0.3), merge_mode='concat'))
model.add(Bidirectional(LSTM(256, return_sequences=True,dropout=0.1, recurrent_dropout
    =0.3), merge_mode='concat'))
model.add(TimeDistributed(Dense(num_tar_tokens, activation='softmax')))
model.compile(loss='categorical_crossentropy', optimizer='rmsprop', metrics=['acc'])
```

bidirectional layer has 256 cells. We use the *softmax* as the activation function of the output layer and the *RMSprop* optimizer in training. We use categorical cross entropy as the loss function, and a batch size of 64 sequences with a sequence length of 400 while wrapping longer sequences similar to our work in [12]. To combine the forward and backward layers in our BiLSTM layers, we use *concatenation*. For the models that use dropout, we use the base settings of dropout = 0.1 and recurrent_dropout = 0.3. We set the training time to a maximum of 50 epochs with early stopping and 5-epoch patience. We show the skeleton code of the 2-Layer model with dropout in Listing 1.

### 4.2 Data Sets

We use two widely used datasets for Arabic NLP [42, 43, 44, 45, 46, 47, 48] to train, validate, and test our proposed BiLSTM models. The first is a processed subset of the Tashkeela corpus as extracted in [49]. The second is from the Linguistic Data Consortium's (LDC) Arabic Treebank (LDC2010T08), specifically speaking: the Arabic Treebank Part 3 (ATB3) v3.2 [50]. We will now-forth refer to this dataset as ATB3 in this paper. The major difference between the Tashkeela and ATB3 datasets is the form of the Arabic text they consist of. Tashkeela mainly contains 55K sequences from classical Arabic texts. On the other hand, ATB3 contains samples of modern standard Arabic of 599 distinct news-wire stories from the Lebanese publication An-Nahar. We show in Table 5 the characteristics of the two datasets in terms of sequence count, word count, character count, average words per sequence, average letters per word, and fraction of sequences shorter than or equal to 400 characters.

Table 5: Tashkeela and ATB3 Datasets Characteristics

| **Criterion** | **Tashkeela** | **ATB3** |
|---|---|---|
| **Sequence count** | 55K | 26K |
| **Word count** | 2,312K | 305K |
| **Character count** | 12,464K | 1,660K |
| **Words per sequence** | 42.1 | 11.3 |
| **Letters per word** | 4.0 | 4.6 |
| **Sequences $\leq$ 400 chars.** | 84.1% | 99.9% |

In Table 6, we present the Arabic letters and their variations that account for the majority of the soft spelling errors that we presented in detail in Section 1 and cited in examples in Table 2. We break them down in order of their absolute frequency within the dataset texts relative to the character count. We also break them down in terms of their relative appearance to each other within the same texts. This table demonstrates that about one fifth of the characters are involved in the common soft spelling mistakes under study and that the relative frequencies of these letters are highly skewed ranging from 0.13% to 58.23%.

We split the ATB3 dataset as proposed by Zitouni *et al.* [51] such that we use the first 509 news-wire stories, in *chronological* order, to train the model and use the remaining 90 stories to validate and test the model. This accounts for 22,170 sequences for training and 3,857 sequences for validation. Similarly, we split the Tashkeela dataset into 50K lines for training, 5,000 lines for testing. In our previous work [12], we analyzed the best maximum sequence length to use based on the same datasets. We found that a maximum sequence of 400 characters provides the best speed versus accuracy trade off. We consequently adopt this sequence length in this work as well.

Table 6: The subset of target Arabic letters (and their variations) under study and their absolute and relative frequencies within the datasets

| | **Frequency** | | **Relative Frequency** | |
|---|---|---|---|---|
| **Letter(s)** | **Tashkeela** | **ATB3** | **Tashkeela** | **ATB3** |
| ا | 8.38% | 11.55% | 46.96% | 58.23% |
| ـه | 2.55% | 0.58% | 14.31% | 2.93% |
| أ | 2.41% | 1.70% | 13.52% | 8.57% |
| ة | 1.22% | 2.49% | 6.81% | 12.55% |
| إ | 0.93% | 0.79% | 5.21% | 3.97% |
| ـو | 0.68% | 0.24% | 3.79% | 1.21% |
| ى | 0.67% | 0.72% | 3.77% | 3.65% |
| ـت | 0.40% | 0.81% | 2.24% | 4.11% |
| اء | 0.18% | 0.24% | 1.03% | 1.24% |
| ئ | 0.18% | 0.40% | 1.03% | 2.04% |
| آ | 0.07% | 0.08% | 0.40% | 0.41% |
| ء | 0.07% | 0.03% | 0.37% | 0.13% |
| ؤ | 0.06% | 0.14% | 0.37% | 0.69% |
| ـوا | 0.04% | 0.06% | 0.34% | 0.28% |
| **Subtotal** | 17.85% | 19.83% | 100.00% | 100.00% |
| **Other chars.** | 82.15% | 80.17% | 0.00% | 0.00% |
| **Total** | 100.00% | 100.00% | 100.00% | 100.00% |

In addition to the above datasets, we test the proposed solutions using samples of real soft spelling mistakes (Test200). These samples were collected in a previous work [8] and are summarized in Table 7. They have a challenging collection of soft spelling mistakes with an average of 6.5 mistakes per sequence.

Table 7: Test200 Test Set of Real Soft Spelling Mistakes

| **Criterion** | **Count** |
|---|---|
| **Sequence count** | 200 |
| **Word count** | 2,443 |
| **Character count** | 24,002 |
| **Number of mistakes** | 1,306 |
| **Mistakes per sequence** | 6.5 |

### 4.3 Training Approaches

We have experimented with the following two approaches to train BiLSTM networks to correct the soft spelling mistakes.

**Transformed Input:** This approach trains the BiLSTM network to predict the correct form of the letters under study given unified transformed input. Once the sequences are converted to the intermediary form, as described in Section 3.3, we transform the letters that are often confused with each other into one final form. We show the used mappings of the letters affected by this transformation in Table 8. All *hamza* forms are transformed to plain *hamza* (ء), the *heh* and *teh* forms at word ends are transformed to *teh marbuta* (ة), *waw-alef* at word ends are transformed to *waw* (ـو), and *alef* at word ends are transformed to *alef maksura* (ى).

**Stochastic Error Injection:** This approach trains the network to correct artificial errors randomly injected in the input sequences. We inject errors in the input sequence by replacing the target letters pertaining to the soft Arabic spelling mistakes. With an *error injection rate p*, a letter belonging to the four groups shown in Table 8 is randomly replaced by one of the letters in its group. For example, we replace *alef maksura* with an *alef* and vice versa (cases 2 and 3 in Table 2). We have investigated using multiple error injection rates $p$ as described in Section 5. For example, 10% of the letters under investigation are replaced with $p = 10\%$.

Table 8: Letters changed in the “Transformed Input” approach

| **Intermediate Forms** | **Mapped To** |
|---|---|
| [ء, آ, أ, ؤ, إ, ئ, ا, and J (اء)] | ء |
| [ ة, T (ـت), and H (ـه)] | ة |
| [ O (ـو) and A (ـوا) ] | O (ـو) |
| [ ا and ى ] | ى |

### 4.4 Evaluation Metrics

We measure and evaluate the performance of BiLSTM networks using their computational time and multiple performance metrics, namely: *accuracy*, *precision*, *recall*, $F_1$ *score*, character error rate ($CER$), and word error rate ($WER$). These metrics are quite common in evaluating the performance of solutions pertaining to error detection and correction, voice recognition, and similar sequence-based problems [13, 36, 49, 52, 53]. The first four measures are readily computed and understood through means of a simple confusion matrix that we will explain within the context of our work. At the character level, any character in any sequence can either be correctly spelled or not. When comparing the predicted outcome of our model with the actual (target) sequence, we can thus have four cases that we show in Fig. 3. For a certain character $c$, the counts of these four cases are:

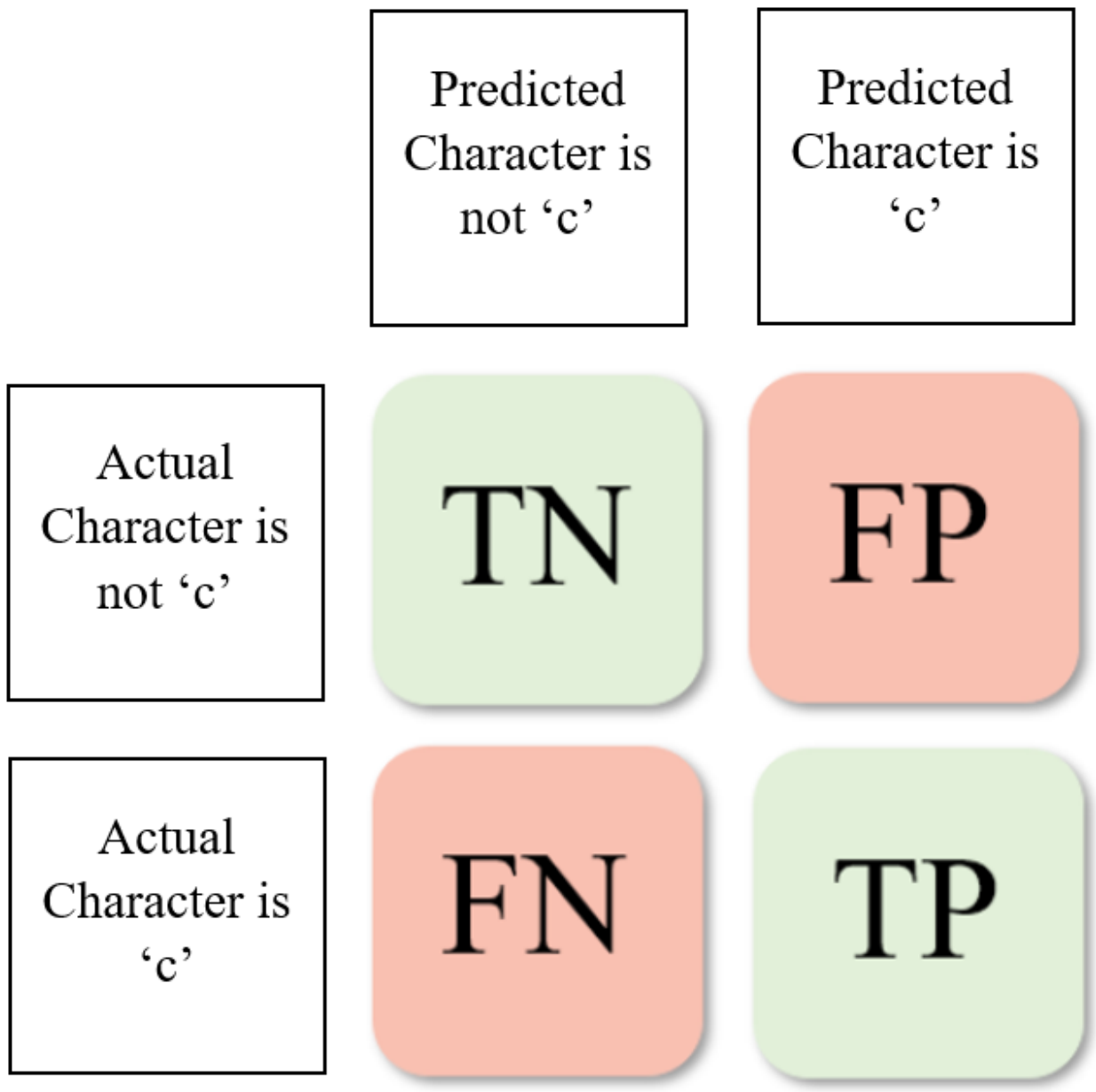

Figure 3: Confusion matrix of correct and erroneous characters between predicted and actual sequences

**True Positive (*TP*):** The number of times character $c$ is correctly predicted as $c$.

**True Negative (*TN*):** The number of times characters other than $c$ are predicted correctly.

**False Positive (*FP*):** The number of times characters other than $c$ are incorrectly predicted as $c$.

**False Negative (*FN*):** The number of times character $c$ is incorrectly predicted as another character.

It is evident that we need our model to maximize the correct cases $TP$ and $TN$ (cases shown in green). We desire our model also to keep the incorrectly predicted characters $FP$ and $FN$ to a minimum (cases shown in red). This readily translates to the definition of the *accuracy* metric that we present in Eq. 8:

$$Accuracy = \frac{TP + TN}{TP + TN + FP + FN} \tag{8}$$

The *precision* metric is the ratio of correct predictions of character $c$ to total number of characters predicted as $c$ and is given by Eq. 9:

$$Precision\ (P) = \frac{TP}{TP + FP} \tag{9}$$

The *recall* metric is the ratio of correct predictions of character $c$ to the total count of actual $c$ characters. Eq. 10 mathematically defines the recall as:

$$Recall\ (R) = \frac{TP}{TP + FN} \tag{10}$$

The $F_1$ score combines the *precision* and *recall* metrics into one score by applying the weighted harmonic mean on both giving equal weights to each. We get the $F_1$ score using Eq. 11:

$$F_1 = 2 \times \frac{P \times R}{P + R} \tag{11}$$

We additionally use a new metric ($FP/Changes$) to assess the prediction error rate with respect to the number of replacements in the input sequence. We divide the number of false positive cases of a letter over the number of times this letter was changed in the input sequence. The accuracy is reported in this work for the entire set of characters. Whereas, precision, recall, $F_1$, and $FP/Changes$ scores are reported as weighted averages of the target letters shown in Table 6.

We also evaluate our models using the character error rate (CER) and word error rate (WER). CER is the percentage of letters that are misspelled whereas WER is the percentage of misspelled words. A word is considered misspelled if it has at least one incorrectly spelled letter.

## 5 Results and Discussion

This section presents the results of evaluating alternative network topologies and the two training approaches proposed above. This section also presents and discusses the detailed performance evaluation results.

### 5.1 Network Topology

We started by evaluating three network topologies: the base network of two BiLSTM layers, another that adds the dropout technique to the base network, and a third that extends the second network to having four BiLSTM layers instead of two. These three alternatives are inspire by experience from previous work, *e.g.*, [41]. In this initial evaluation, we use the *transformed input* training approach. For each of these three networks, we use the Tashkeela and ATB3 sets to separately train two separate instances of the BiLSTM network; one trained for classical Arabic, and another for MSA Arabic. We repeat this procedure for the three networks using the same training sequences and compare their performance in terms of *accuracy*, *precision*, *recall*, and $F_1$ score on the test set.

In Fig. 4, we show these metrics for the networks trained with the Tashkeela set. Despite the results being marginally close, we concur that networks with the *dropout* technique employed perform generally better than the one without. We observe that there is no large difference between the performance of the 2-layer and 4-layer networks with dropout, with only a slight edge towards the 2-layer network. Fig. 5 similarly illustrates the same metrics but for the case when we evaluate the BiLSTM networks using the ATB3 set. We notice that our observations and conclusions for the Tashkeela set carry over to the ATB3 set. However, the accuracy with the ATB3 set is generally lower than the accuracy on the larger Tashkeela set.

We notice also that the accuracy is higher than the other three metrics because it is calculated for the entire character set while the other metrics are weighted averages of the target letters only. For the remaining experiments, and given the slight performance edge and lesser complexity of the 2-layer BiLSTM network with dropout used, we conduct and evaluate the remaining experiments and report their results based on this model only.

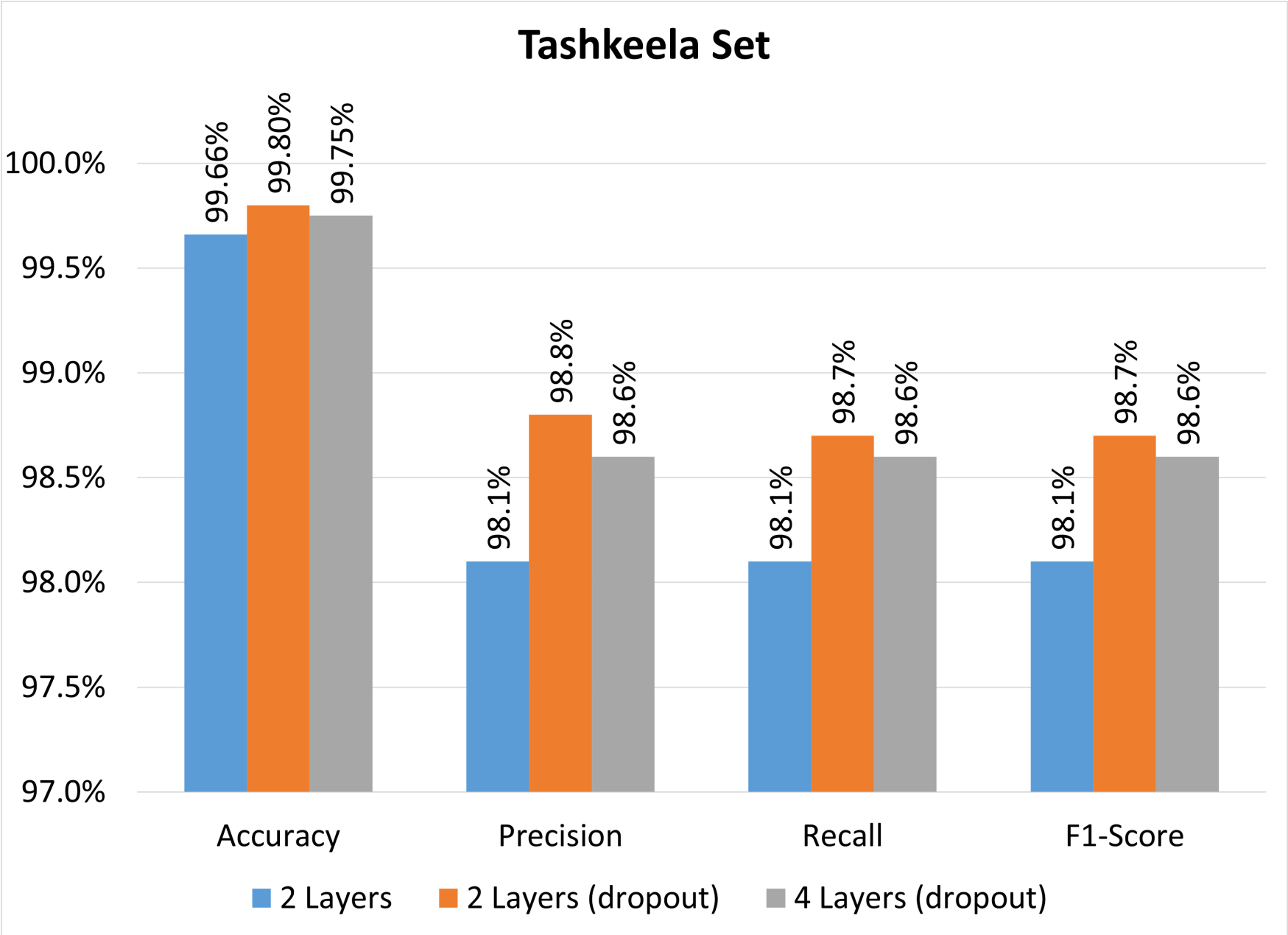

Figure 4: Performance of three BiLSTM networks using the transformed input training on the "Tashkeela" set.

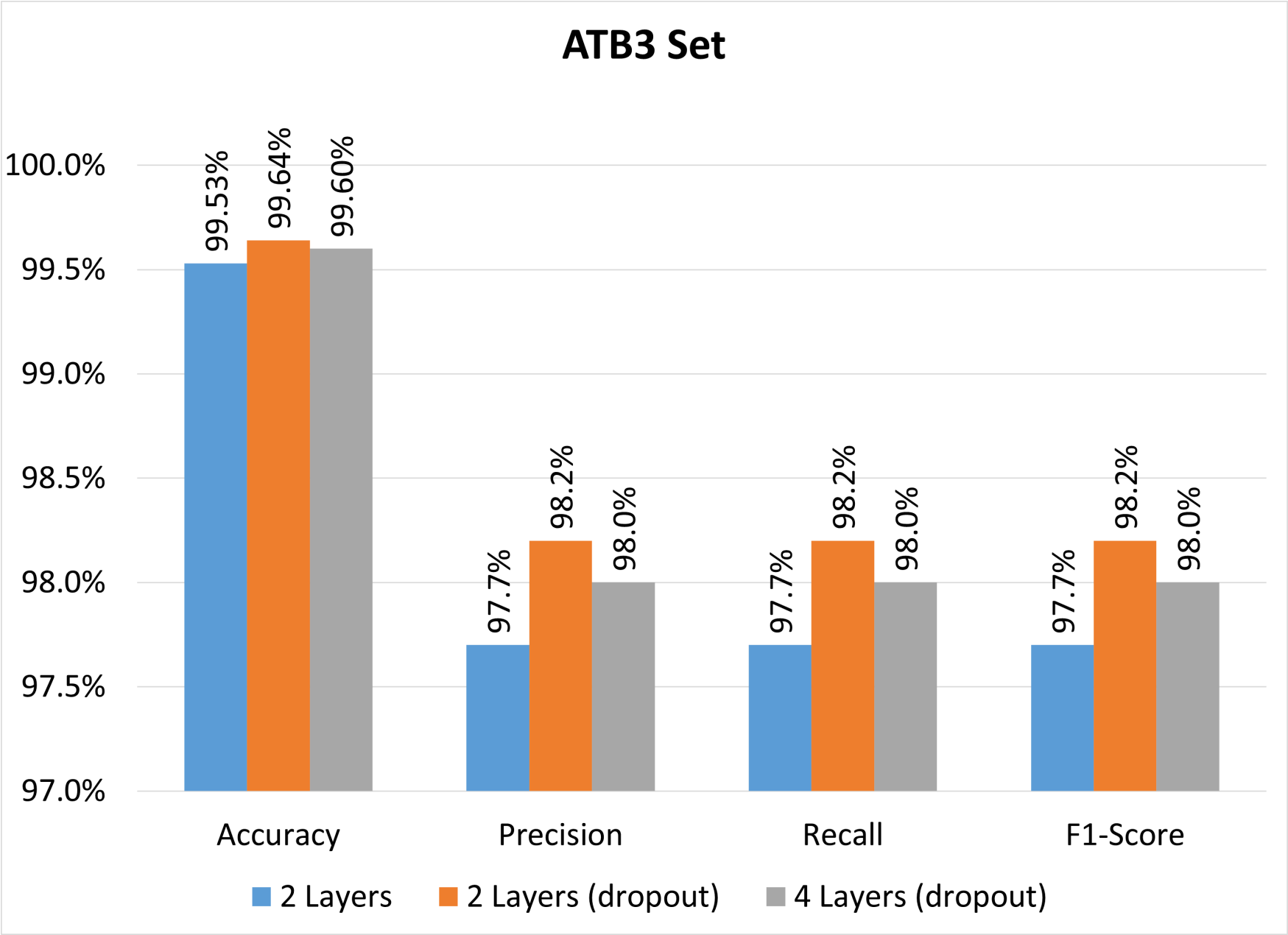

Figure 5: Performance of three BiLSTM networks using the transformed input training on the "ATB3" set.

### 5.2 Training Approach

In this section, we present the performance of the adopted 2-layer network on the two proposed training approaches: *transformed input* and *stochastic error injection*. The error injection rate $p$ correlates with the network's ability to correct errors. Therefore, we experimented with multiple rates, *e.g.*, 2.5%, 10%, and 40%, alongside the transformed input approach.

For this evaluation, we show the results also in terms of *accuracy*, *precision*, *recall*, and $F_1$ score. Recall that these metrics are found not only from the letters that we actually changed, but also include the whole letters in the subset under study. Therefore, we expect that the inclusion of all the letters that could possibly be changed and those actually changed dilutes the results. For example, for the case when the error injection rate is 2.5%, we already know that a high percentage of the remaining 97.5% letters are correct and match the target output. Despite the expected result dilution, we have to present these results for in most cases the analysis and the interest is in the overall output character sequence and that it should be error free. Given these disproportional ratios between unchanged letters and error-injected letters, we expect better performance with lower error injection rates. We indeed observe these results in Fig. 6 and Fig. 7. The best performance appears here for the network with the smallest error injection rate of $p = 2.5\%$.

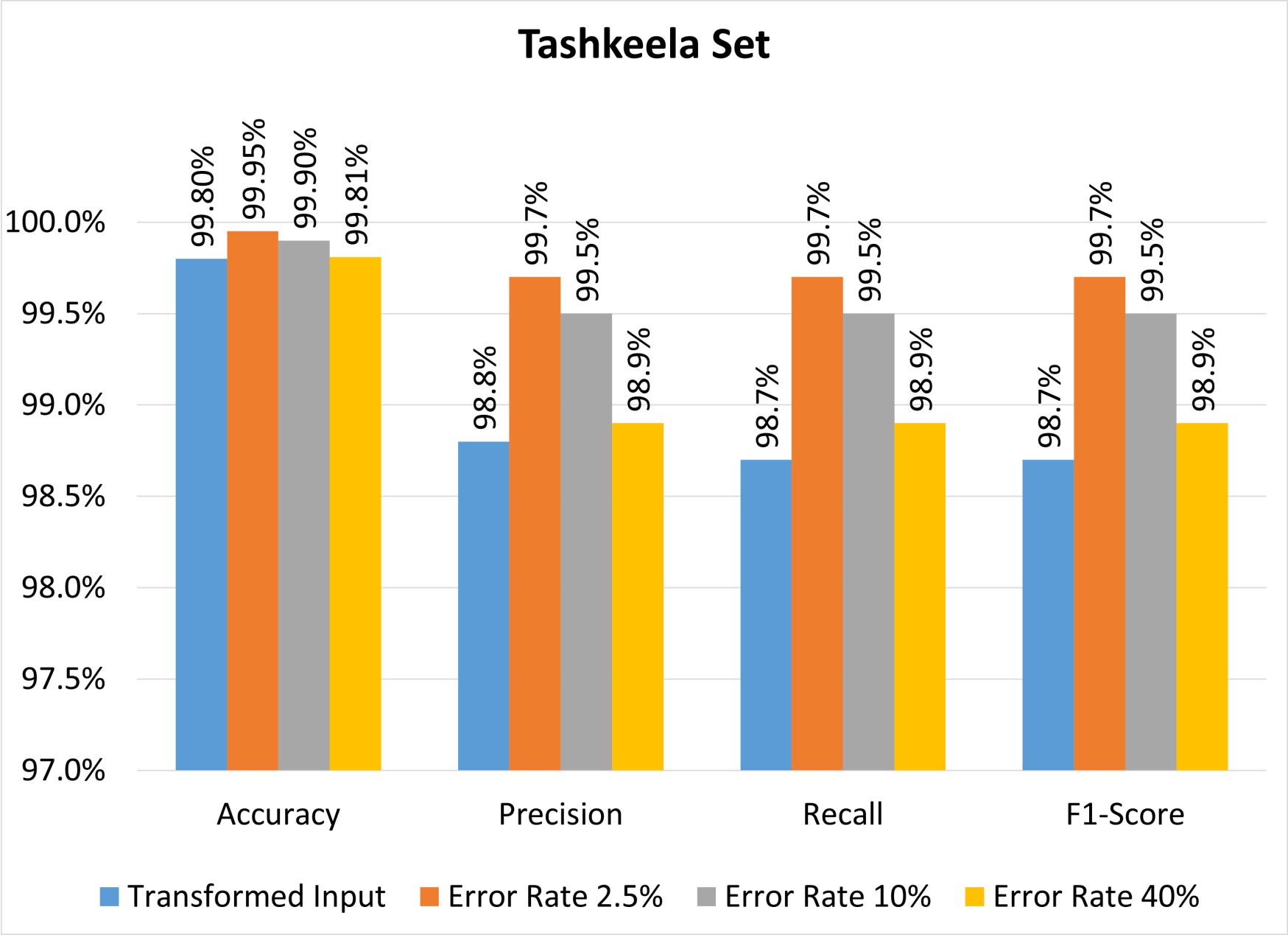

Figure 6: Performance of the transformed input and stochastic error injection training approaches on the "Tashkeela" set using the 2-layer with dropout network. Three error injection rates are investigated.

It is difficult to make solid conclusions about the two training approaches from the previous data alone. So we compare the two training approaches using the *FP*/*Changes* ratio. This allows us to observe more insights that we could not deduce from the diluted results in Fig. 6 and Fig. 7. We present this ratio in Fig. 8 for the Tashkeela and ATB3 sets. In this analysis, we easily see that increasing the error injection rate helps train the network to better detect and correct errors. For example for Tashkeela, we observe a decrease of the percentage of false positives from 14.4% to 3.6% as we increase the error injection rate from 2.5% to 40%. Despite the more errors introduced, the network was able to handle them well. In the case of transformed input, we even observe better performance. This is because transforming the most confused characters into one form is another way of introducing different errors into the network training.

In Fig. 9 and Fig. 10, we show the character and word error rates for the proposed training approaches. These two metrics are global metrics similar to the accuracy. Therefore, they are generally low because most characters in the input are correct and only need to be passed to the model output as is. We present these metric as any application will handle entire sequences with possible spelling mistakes and attempt to provide an error-free version. Essentially, CER and WER are related so we expect a similar pattern for both. We expect this pattern to resemble that for the analysis presented in above, specifically, $1-$ *accuracy*. This is due to the ratio of already correct and unchanged characters that are in the input, predicted, and target sequences. Despite this, we observe that for both experiments on the Tashkeela and ATB3 datasets that in the worst case the CER did not exceed 0.36% while the WER did not exceed 1.88%. Referring to the analysis of the datasets that we show in Table 5, we note that the average number of letters per word is 4.0 and 4.6 for the Tashkeela and ATB3 sets, respectively. The results we show for WER in Fig. 10 are approximately five times than those for CER in Fig. 9. This is in line with the letters per word statistic for these datasets under consideration.

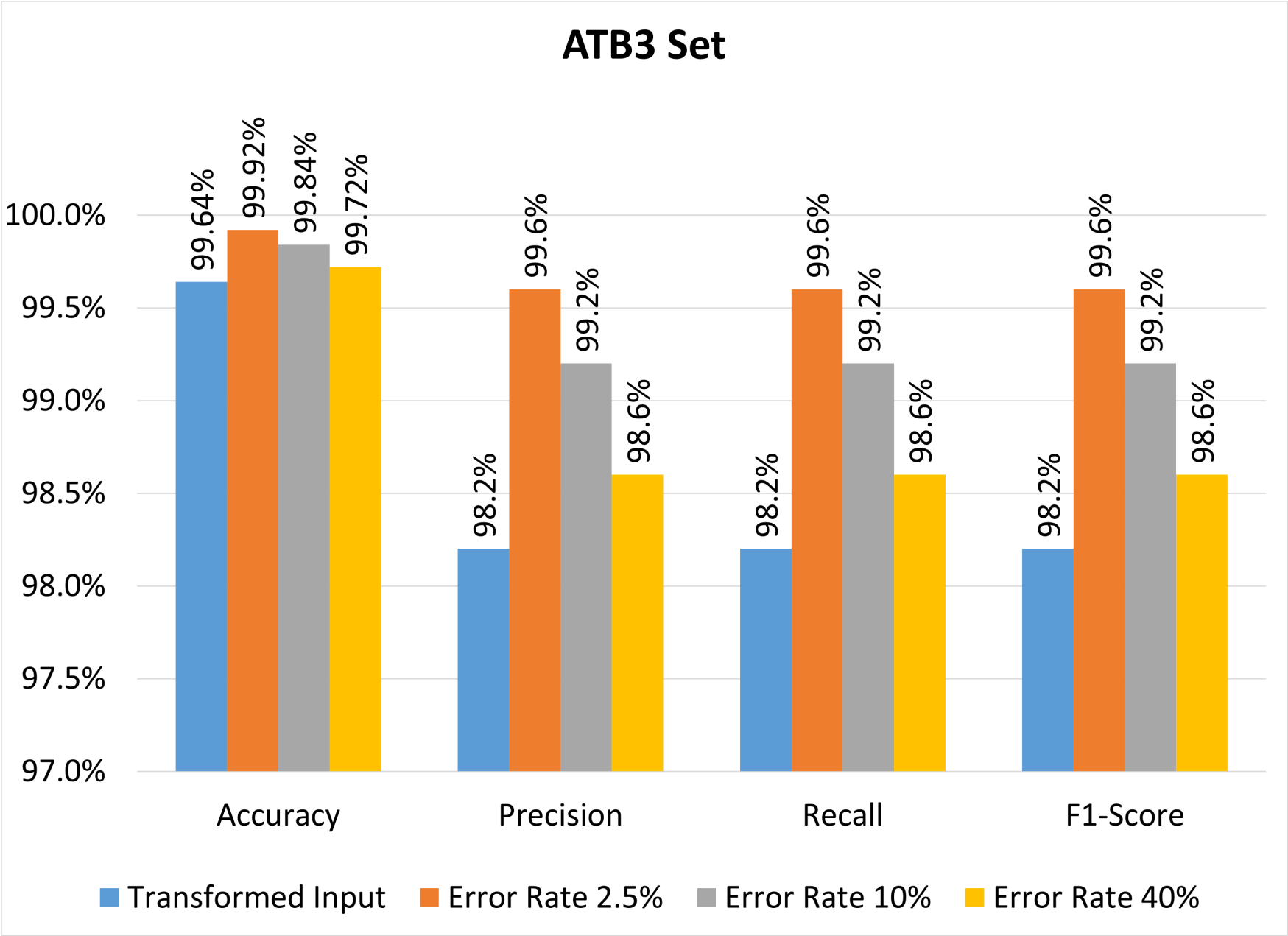

Figure 7: Performance of the transformed input and stochastic error injection training approaches on the "ATB3" set using the 2-layer with dropout network. Three error injection rates are investigated.

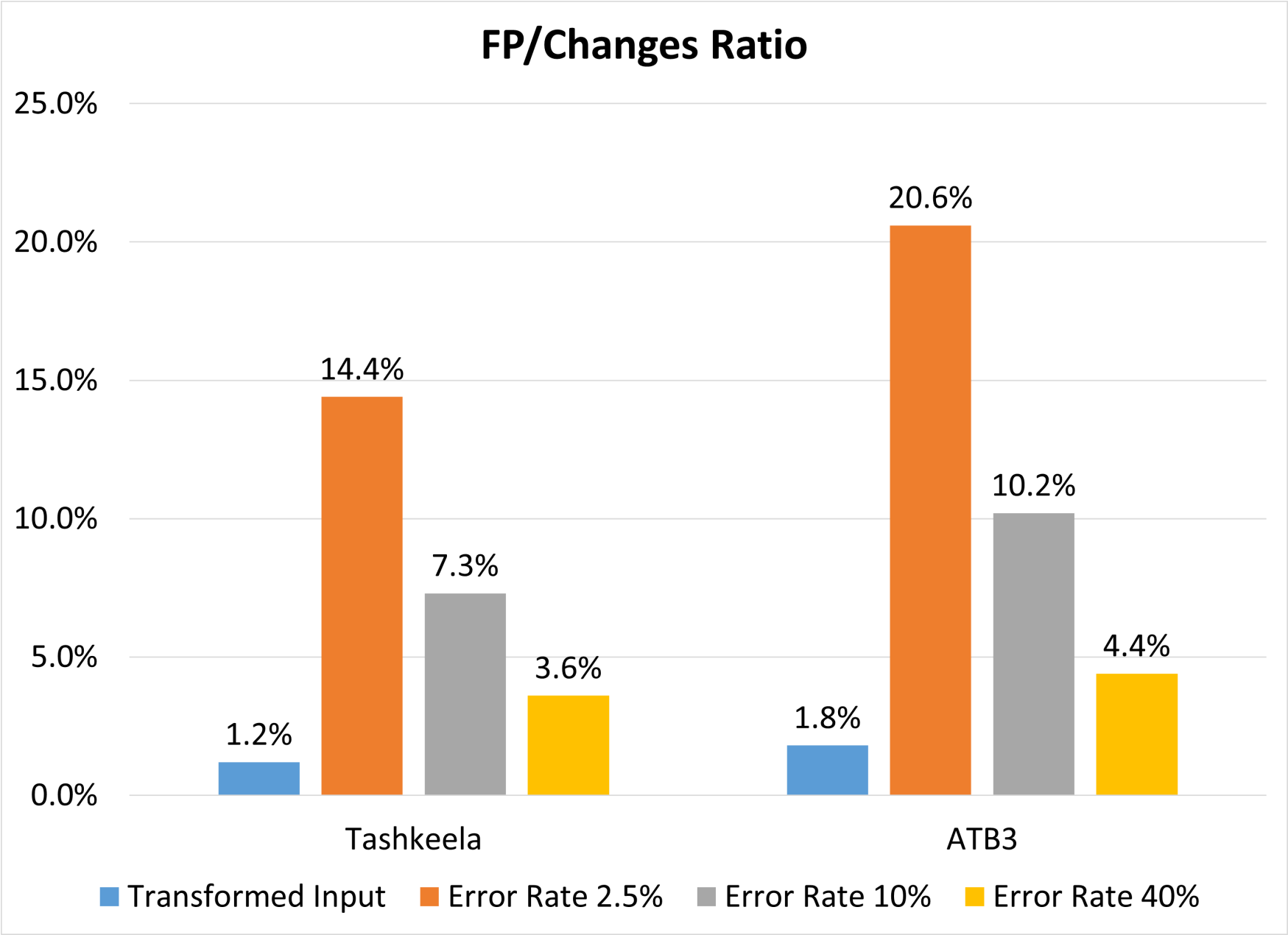

Figure 8: The *FP*/*Changes* ratio of the two training approaches on the Tashkeela and ATB3 sets.

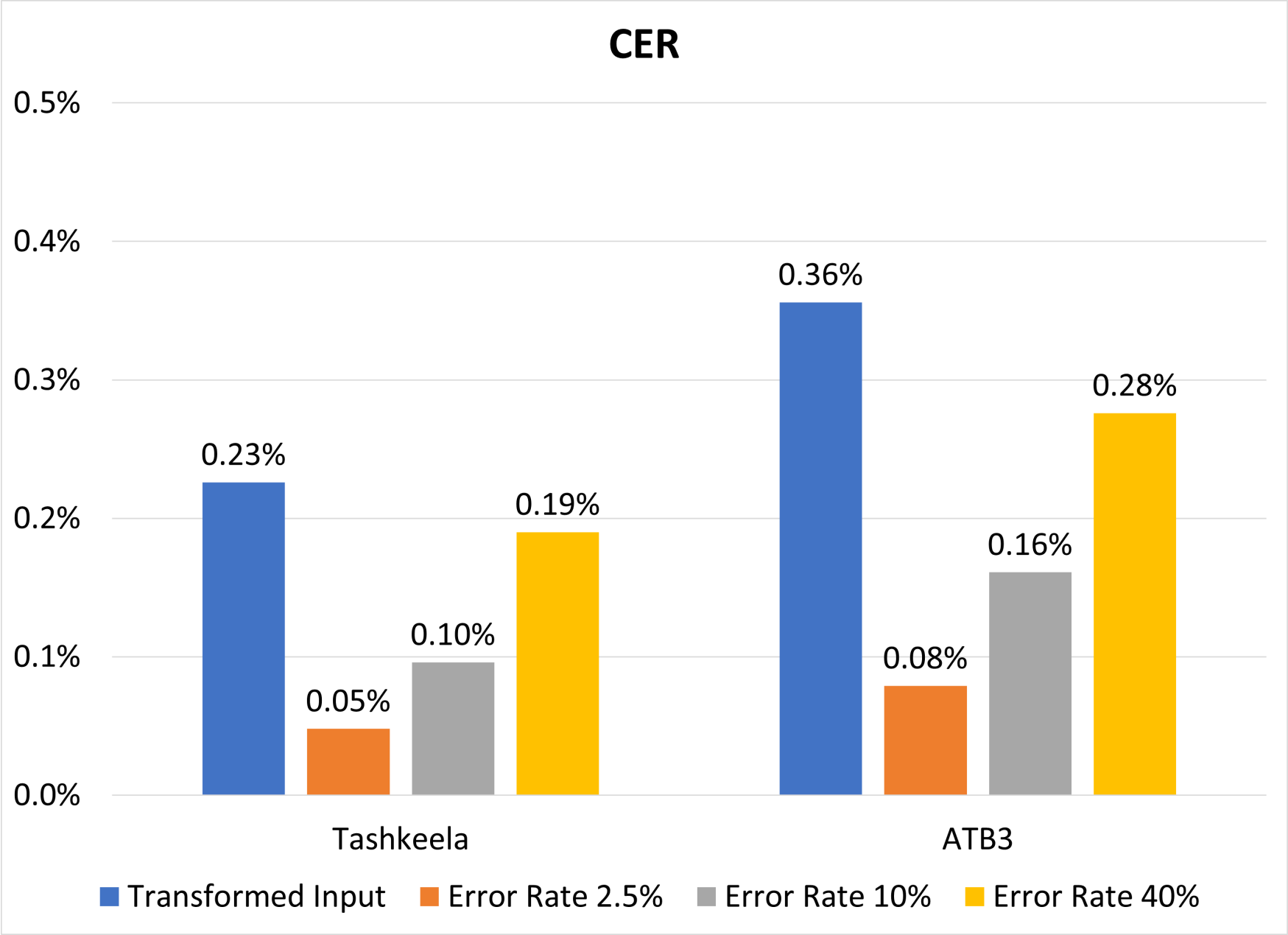

Figure 9: Character error rate of the two training approaches on the Tashkeela and ATB3 sets.

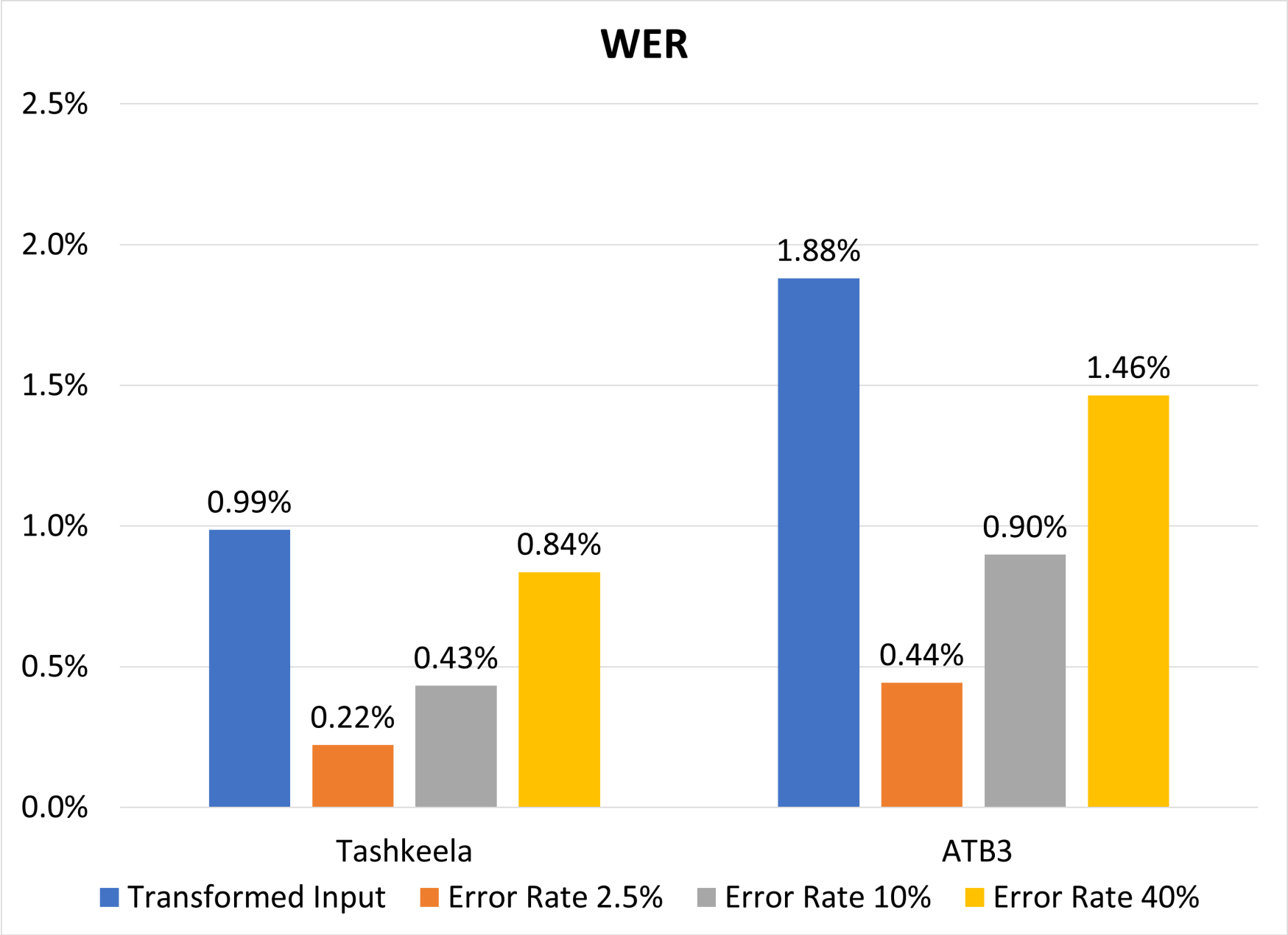

Figure 10: Word error rate of the two training approaches on the Tashkeela and ATB3 sets.

### 5.3 Test200 Results

The above results are not conclusive about which training approach is best. Therefore, we used the BiLSTM models trained on the two training approaches to correct the mistakes in the Test200 set. We use the networks which we trained using the Tashkeela and ATB3 sets here. We report the CER of the predicted Test200 sequences on the eight training configurations shown in Fig. 11. Except for the 2.5% error injection rate, the two training sets provide results within 0.2% for each other. Yet, we note that despite that the reported CER for this external set is quite low, it is an order of magnitude higher than that reported for the test sequences of the Tashkeela and ATB3 datasets. This is because the BiLSTM models were trained using artificially injected errors that might not necessarily always correspond to errors committed by real-users. This stresses the need for a large Arabic corpora collected from real users, annotated, and corrected by linguistic experts to enrich the Arabic NLP research domain.

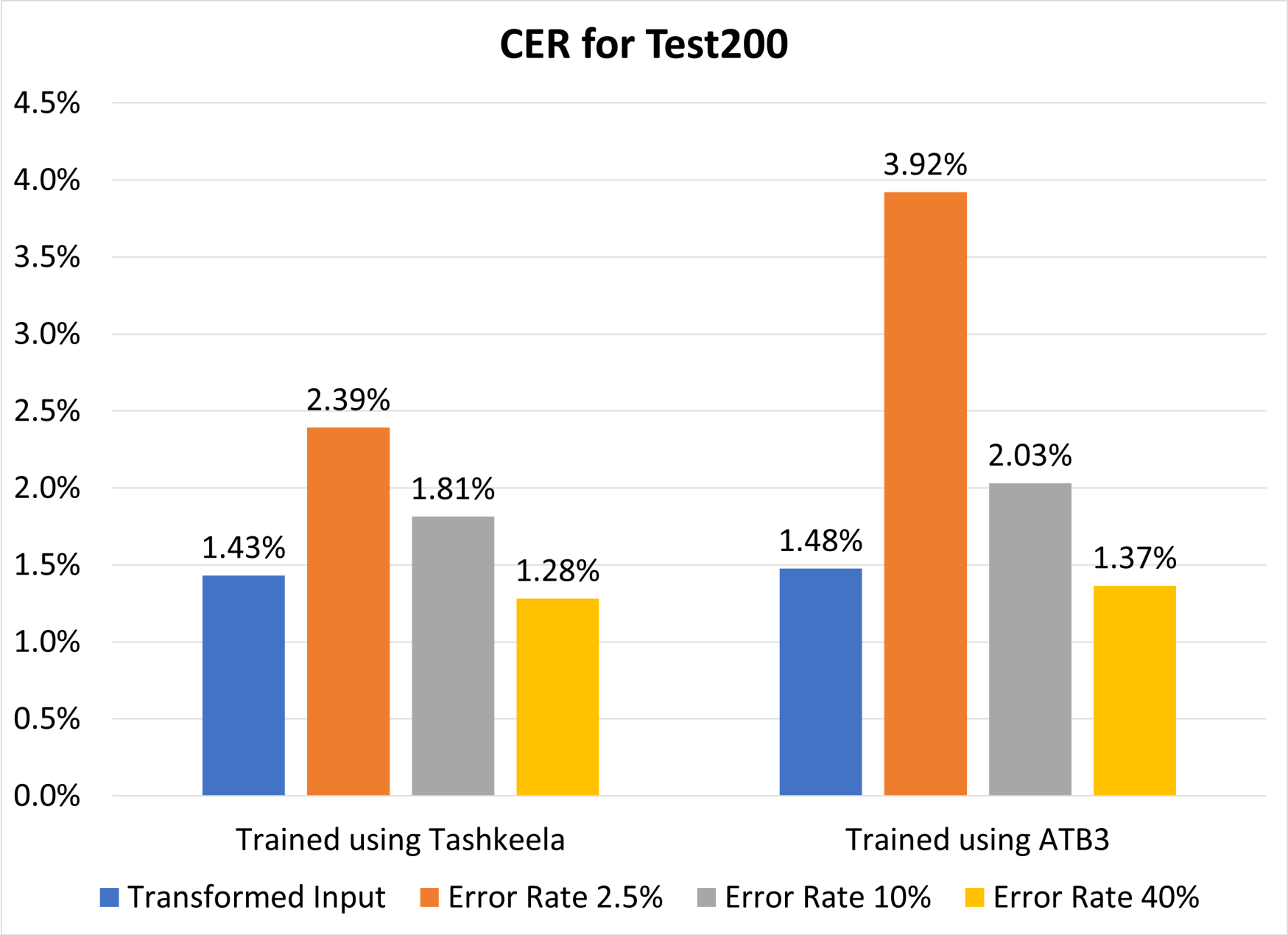

Figure 11: CER of the two training approaches on the Test200 set for models trained on the Tashkeela and ATB3 sets.

Figure 11 shows also that training on the larger Tashkeela set gives better results and the transformed input and error injection with 40% rate are better than lower rates. The best results with only 1.28% CER is for the case when training using Tashkeela set and 40% error injection rate. To test whether increasing the error injection rate beyond 40% would further decrease CER, we experimented with higher rates (50%, 70%, and 100%). Figure. 12 shows that the model with $p = 40\%$ performs best and yielded the least character error rate. Therefore, we recommend using error injection training approach with $p = 40\%$. We further analyze the results of this approach in the following subsection.

### 5.4 Confusion Matrix Results

Figure 13 shows the confusion matrix for the predicted against actual of the Tashkeela test set of the letters under investigation. We present here this matrix for the best trained model (error injection with $p = 40\%$) and Tashkeela set because it is the largest set we use. This matrix allows us to analyze the letters that our machine learning model mostly confuses with each other. While the majority of the letters are correctly classified and corrected (diagonal), most confusion occurs within the three terminal letters: *heh* (ـه), *teh* (ـت), and *teh marbuta* (ة), and *alef with hamza above* (أ) and *alef with hamza below* (إ). We list Examples 5, 12, and 16 of these common mistakes in Table 2. For example, out of $(151 + 46 + 6,881 = 7,078)$ *teh marbuta* letters, 151 and 46 are wrongly predicted as *heh* and *teh*, respectively.

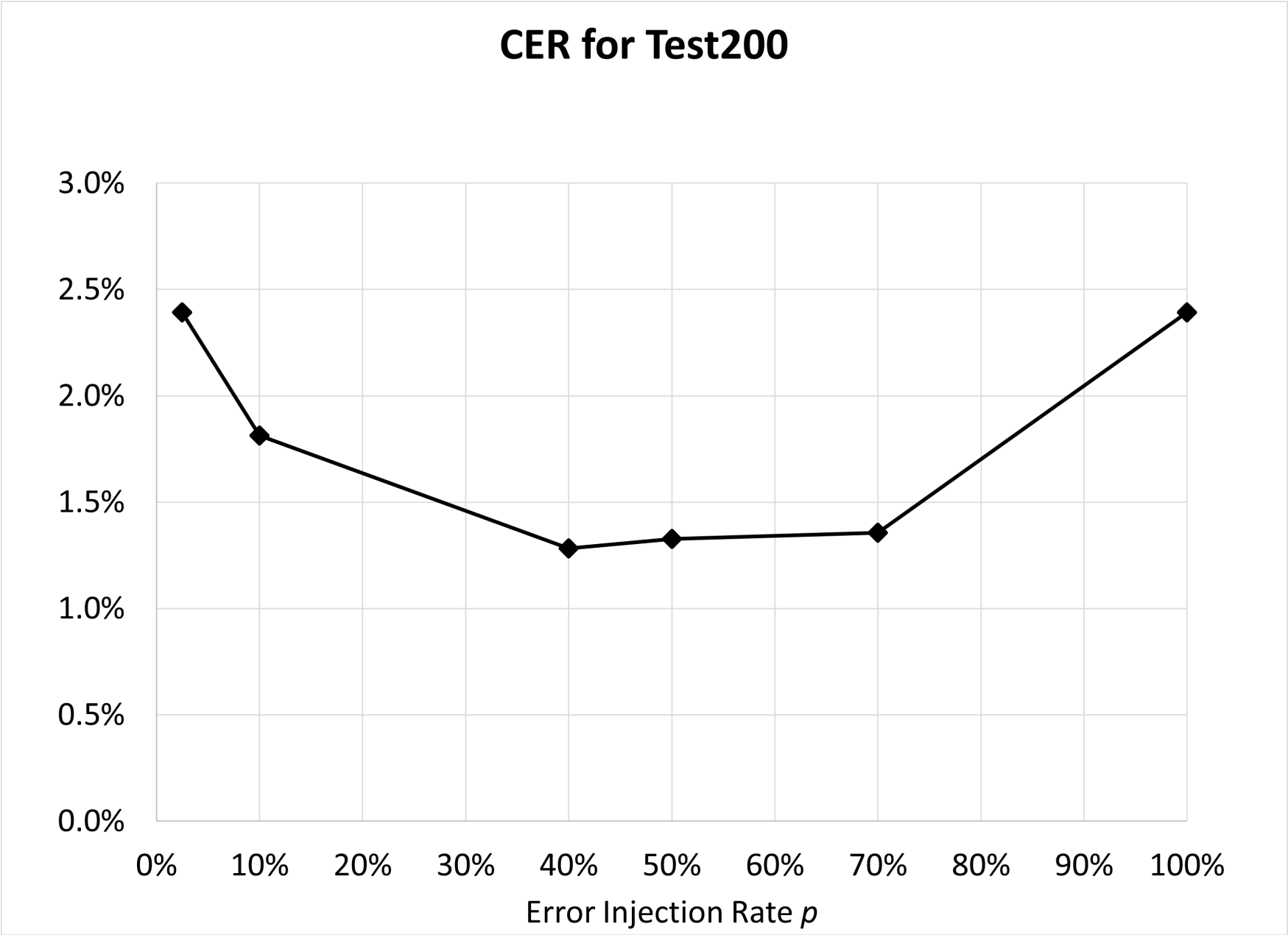

Figure 12: CER of Test200 set for models trained with the stochastic error injection approach and the Tashkeela set as a function of the error injection rate $p$.

| Actual Letter \ Predicted Letter | وا | ه | اء | و | ت | ء | آ | أ | ؤ | إ | ئ | ا | ة | ى |
|---|---|---|---|---|---|---|---|---|---|---|---|---|---|---|
| وا | 264 | 0 | 0 | 3 | 0 | 0 | 0 | 0 | 0 | 0 | 0 | 0 | 0 | 0 |
| ه | 0 | 14,699 | 0 | 0 | 66 | 0 | 0 | 0 | 0 | 0 | 0 | 0 | 102 | 0 |
| اء | 0 | 0 | 1,083 | 0 | 0 | 0 | 0 | 3 | 0 | 0 | 2 | 7 | 0 | 0 |
| و | 9 | 0 | 0 | 3,888 | 0 | 0 | 0 | 0 | 0 | 0 | 0 | 0 | 0 | 0 |
| ت | 0 | 123 | 0 | 0 | 2,242 | 0 | 0 | 0 | 0 | 0 | 0 | 0 | 30 | 0 |
| ء | 0 | 0 | 0 | 0 | 0 | 381 | 0 | 0 | 0 | 0 | 0 | 0 | 0 | 0 |
| آ | 0 | 0 | 0 | 0 | 0 | 0 | 359 | 25 | 0 | 2 | 0 | 2 | 0 | 0 |
| أ | 0 | 0 | 3 | 0 | 0 | 1 | 20 | 13,660 | 19 | 79 | 6 | 31 | 0 | 0 |
| ؤ | 0 | 0 | 0 | 0 | 0 | 0 | 0 | 15 | 315 | 0 | 17 | 4 | 0 | 0 |
| إ | 0 | 0 | 0 | 0 | 0 | 0 | 6 | 75 | 0 | 5,358 | 1 | 3 | 0 | 0 |
| ئ | 0 | 0 | 2 | 0 | 0 | 2 | 0 | 4 | 5 | 0 | 957 | 4 | 0 | 0 |
| ا | 0 | 0 | 2 | 0 | 0 | 1 | 6 | 63 | 1 | 3 | 3 | 48,505 | 0 | 40 |
| ة | 0 | 151 | 0 | 0 | 46 | 0 | 0 | 0 | 0 | 0 | 0 | 0 | 6,881 | 0 |
| ى | 0 | 0 | 0 | 0 | 0 | 0 | 0 | 0 | 0 | 0 | 0 | 107 | 0 | 3,735 |

Figure 13: Confusion matrix for the 40% error injection training approach on Tashkeela test set.

### 5.5 Model Timing

Finally, we report the timing metrics (training time and number of training epochs) of six selected configurations in Figures 14 and 15 for Tashkeela and ATB3, respectively. In all cases, the training time for the Tashkeela set is between two to five times more than that for the ATB3. It took the longest to train the 4-layer network for both sets; 48.5 hours and 9.2 hours for Tashkeela and ATB3 using the transformed input approach, respectively. When training the network using variable error injection rate, we observe that the training time *generally* increases as this rate increases.

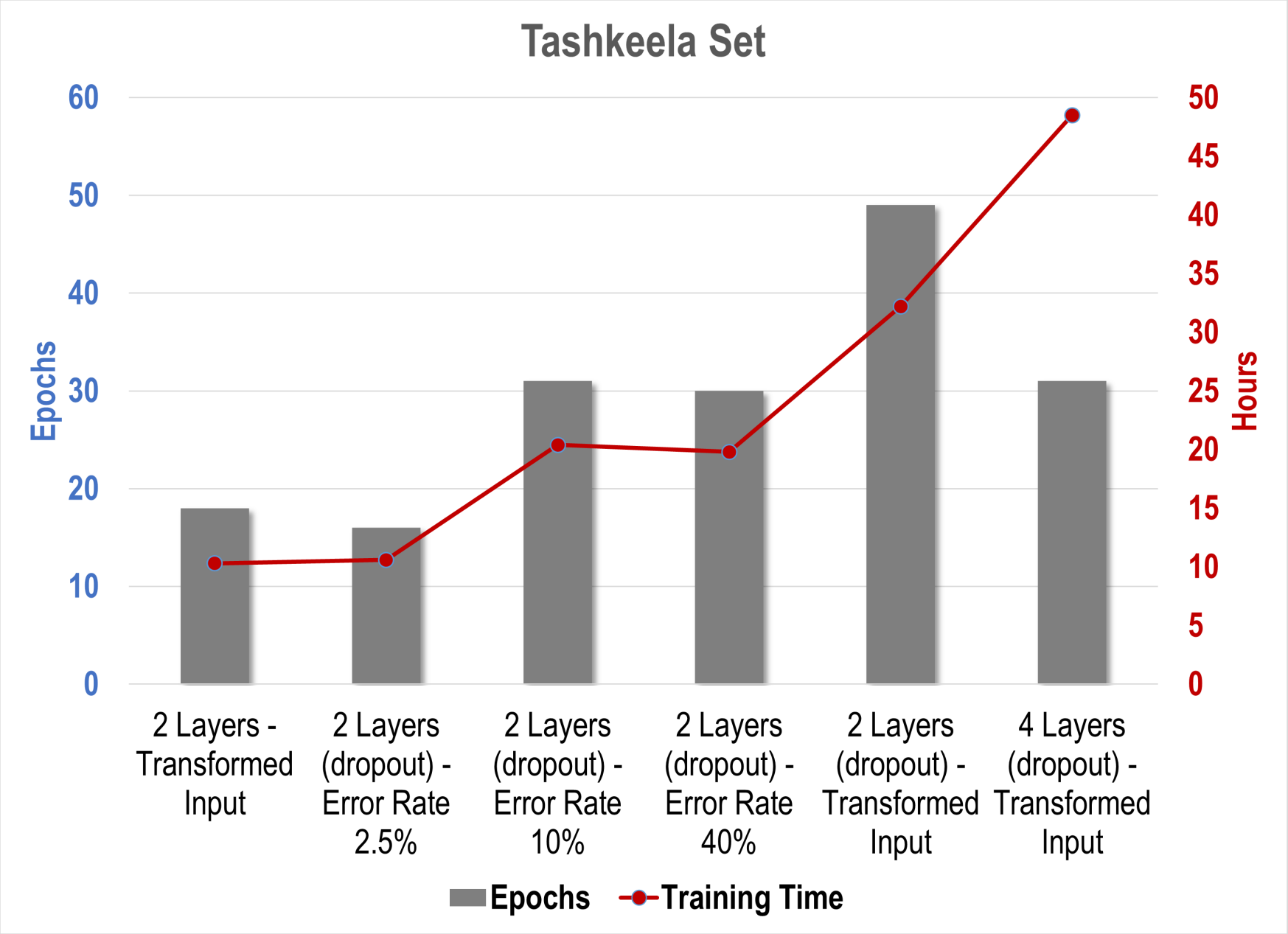

Figure 14: BiLSTM timing metrics under different configurations for the Tashkeela set.

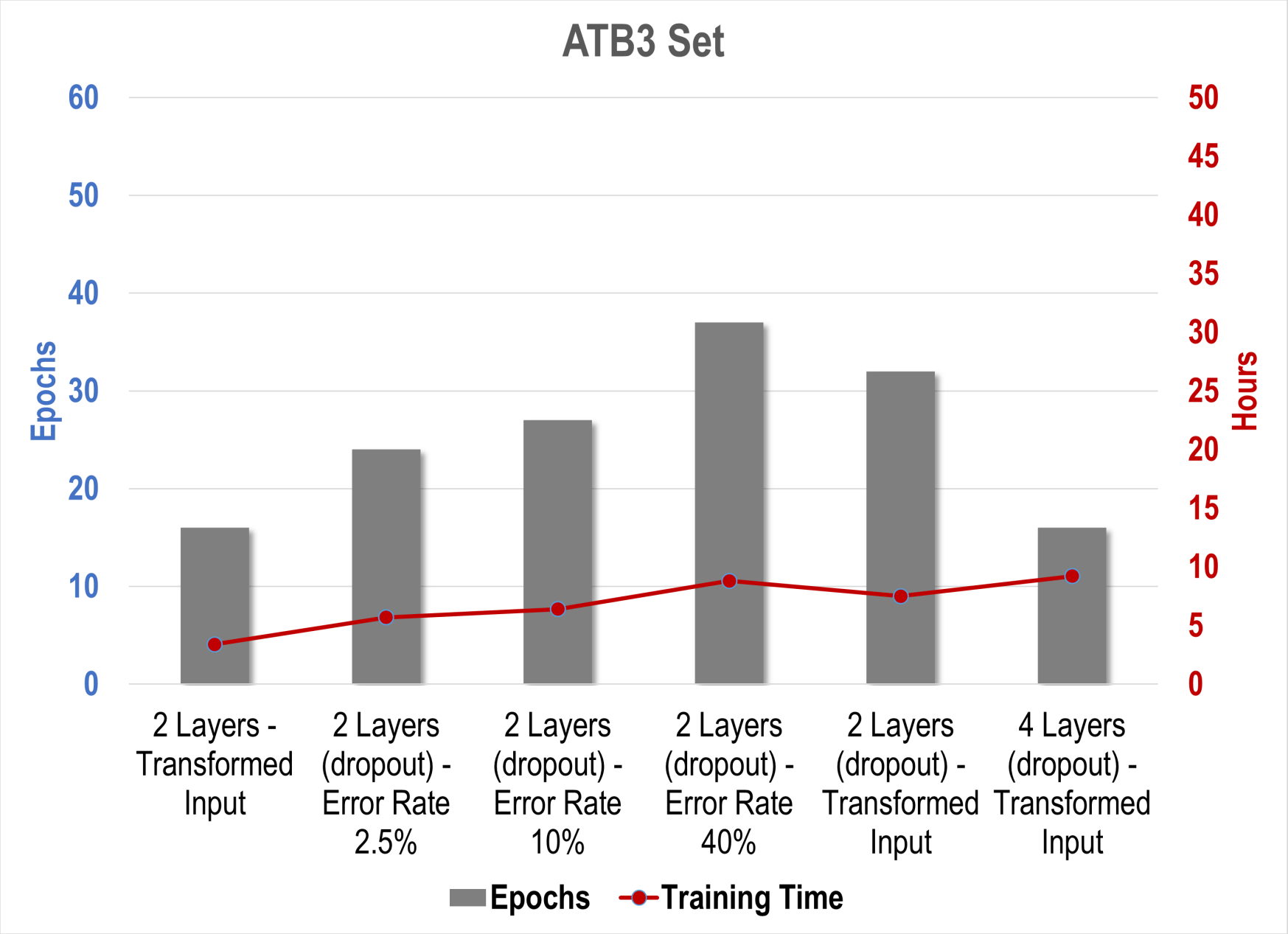

Figure 15: BiLSTM timing metrics under different configurations for the ATB3 set.

In these experiments, we set the maximum number of training epochs to 100. Yet, all models converged before half the set number of epochs. In contrast to the number of hours parameter, we observe that in some cases it took from $\frac{2}{3}$ to twice the number of epochs to train the same model for the two datasets.

## 6 Conclusion

In this work, we addressed the problem of correcting common soft Arabic spelling errors. We developed variant configurations of bidirectional LSTM networks with either two or four hidden layers, while using or forgoing the dropout technique. We use sequences of Arabic texts that are either written in classical Arabic (Tashkeela set), or MSA Arabic (ATB3 set) to train and validate our models. The 2-layer network with dropout had the highest $F_1$ score of 98.7% on Tashkeela test set using the transformed input training approach.

We also experimented with a second training approach where we introduce stochastic errors and train the BiLSTM network to correct them. We deliberately varied the percentage of injected errors in the input sequence to assess the BiLSTM network performance and sensitivity given how well it can learn from lower or higher injected error rates. Networks trained with higher error rates (from 2.5%, 10%, through 40%) have worse accuracy, precision, recall, $F_1$ score, CER, and WER. However, they are more capable of correcting errors as reflected by their *FP/Changes* ratio. Noting that injecting more errors in the input stream beyond 40% does not improve the network performance in correcting soft spelling mistakes.

The transformed input training approach has better *FP/Changes* ratio than that of the stochastic error injection approach. However, the latter approach is better in correcting the soft spelling mistakes in the Test200 test set with error injection rate of $p = 40\%$. The best result on this test set is an CER of 1.28%. This lowest error rate is for the network trained on Tashkeela set and $p = 40\%$.

For future work, we consider expanding beyond the class of soft Arabic spelling errors. Further, a much larger annotated and pre-processed dataset will open doors to improving accuracy. We could possibly handle Arabic spelling correction and letter diacritization in one problem space.

**Acknowledgments** The authors would like to thank the University of Jordan and its Deanship of Scientific Research for supporting this work.

## References

[1] UNESCO. World Arabic language day, Dec 2019.

[2] Ethnologue. Arabic language statistics, 2020.

[3] J. Clement. Internet: Most common languages online 2019, Jul 2019.

[4] Kareem Darwish and Walid Magdy. Arabic information retrieval. *Foundations and Trends in Information Retrieval*, 7(4):239–342, 2014.

[5] Teresa Tinsley and Kathryn Board. Languages for the future. Report ISBN 978-0-86355-883-2, British Council, 2017.

[6] Glosbe. Transliteration and Romanization utilities, 2020.

[7] Karin C Ryding. Learning Arabic as a foreign language, 2020.

[8] G Abandah, M Khedher, W Anati, A Zghoul, S Ababneh, and Mamoun S Hattab. The Arabic language status in the Jordanian social networking and mobile phone communications. In *7th Int'l Conference on Information Technology (ICIT 2015)*, pages 449–456, 2015.

[9] The school of language studies - language categories, 2017.

[10] Gheith A Abandah, Alex Graves, Balkees Al-Shagoor, Alaa Arabiyat, Fuad Jamour, and Majid Al-Taee. Automatic diacritization of Arabic text using recurrent neural networks. *International Journal on Document Analysis and Recognition (IJDAR)*, 18(2):183–197, 2015.

[11] S. Alqudah, G. Abandah, and A. Arabiyat. Investigating hybrid approaches for Arabic text diacritization with recurrent neural networks. In *2017 IEEE Jordan Conference on Applied Electrical Engineering and Computing Technologies (AEECT)*, pages 1–6, Oct 2017.

[12] Gheith Abandah and Asma Abdel-Karim. Accurate and fast recurrent neural network solution for the automatic diacritization of Arabic text. *Jordanian Journal of Computers and Information Technology*, 6(2):103–121, 2020.

[13] Aqil M Azmi, Manal N Almutery, and Hatim A Aboalsamh. Real-word errors in Arabic texts: A better algorithm for detection and correction. *IEEE/ACM Transactions on Audio, Speech, and Language Processing*, 27(8):1308–1320, 2019.

[14] Asma Zahui, Wahiba Elhor, and Mohamed Amine Cheragui. EL-Mossahih V1.0: A hybrid approach for detection and correction of typographical and phonetic transcription errors in Arabic texts. In *2017 8th International Conference on Information Technology (ICIT)*, pages 774–779. IEEE, 2017.

[15] Mohammed Attia, Pavel Pecina, Younes Samih, Khaled Shaalan, and Josef Van Genabith. Arabic spelling error detection and correction. *Natural Language Engineering*, 22(5):751–773, September 2016.

[16] Abd-Muhsin Hamad Al-Ameri. Common spelling mistakes among students of teacher education institutes. *The Islamic College University Journal*, 1(33):445–474, 2015.

[17] Michael Flor, Michael Fried, and Alla Rozovskaya. A benchmark corpus of English misspellings and a minimally-supervised model for spelling correction. In *Proceedings of the Fourteenth Workshop on Innovative Use of NLP for Building Educational Applications*, pages 76–86, Florence, Italy, 2019. Association for Computational Linguistics.

[18] Migucl BIazquez and Catherine Fan. The efficacy of spell check packages specifically designed for second language learners of Spanish. *Pertanika Journal of Social Sciences & Humanities*, 27(2), 2019.

[19] Michael Flor, Yoko Futagi, Melissa Lopez, and Matthew Mulholland. Patterns of misspellings in L2 and L1 English: a view from the ETS spelling corpus. *Bergen Language and Linguistics Studies*, 6, 2015.

[20] Hao Li, Yang Wang, Xinyu Liu, Zhichao Sheng, and Si Wei. Spelling error correction using a nested RNN model and pseudo training data. *arXiv:1811.00238 [cs]*, November 2018.

[21] Eva D'hondt, Cyril Grouin, and Brigitte Grau. Low-resource OCR error detection and correction in French clinical texts. In *Proceedings of the Seventh International Workshop on Health Text Mining and Information Analysis*, pages 61–68, Auxtin, TX, 2016. Association for Computational Linguistics.

[22] Eva D'hondt, Cyril Grouin, and Brigitte Grau. Generating a training corpus for OCR post-correction using encoder-decoder model. In *Proceedings of the Eighth International Joint Conference on Natural Language Processing*, pages 1006–1014, 2017.

[23] Mohammad Bagher Dastgheib, Seyed Mostafa Fakhrahmad, and Mansoor Zolghadri Jahromi. Perspell: A New Persian Semantic-based Spelling Correction System. *Digital Scholarship in the Humanities*, 32(3):543–553, 03 2016.

[24] Azita Yazdani, Marjan Ghazisaeedi, Nasrin Ahmadinejad, Masoumeh Giti, Habibe Amjadi, and Azin Nahvijou. Automated misspelling detection and correction in Persian clinical text. *Journal of Digital Imaging*, pages 1–8, 2019.

[25] Shahin Salavati and Sina Ahmadi. Building a lemmatizer and a spell-checker for Sorani Kurdish. *CoRR*, abs/1809.10763, 2018.

[26] Pravallika Etoori, Manoj Chinnakotla, and Radhika Mamidi. Automatic spelling correction for resource-scarce languages using deep learning. In *Proceedings of ACL 2018, Student Research Workshop*, pages 146–152, Melbourne, Australia, 2018. Association for Computational Linguistics.

[27] Mourad Mars. Toward a robust spell checker for Arabic text. In *International Conference on Computational Science and Its Applications*, pages 312–322. Springer, 2016.

[28] Nouf AlShenaifi, Rehab AlNefie, Maha Al-Yahya, and Hend Al-Khalifa. ARIB@ QALB-2015 shared task: a hybrid cascade model for Arabic spelling error detection and correction. In *Proceedings of the Second Workshop on Arabic Natural Language Processing*, pages 127–132, 2015.

[29] Hamdy Mubarak and Kareem Darwish. Automatic correction of arabic text: A cascaded approach. In *Proceedings of the EMNLP 2014 Workshop on Arabic Natural Language Processing (ANLP)*, pages 132–136, 2014.

[30] Houda Bouamor, Hassan Sajjad, Nadir Durrani, and Kemal Oflazer. Qcmuq@ qalb-2015 shared task: Combining character level mt and error-tolerant finite-state recognition for Arabic spelling correction. In *Proceedings of the Second Workshop on Arabic Natural Language Processing*, pages 144–149, 2015.

[31] Hatem M Noaman, Shahenda S Sarhan, and M Rashwan. Automatic Arabic spelling errors detection and correction based on confusion matrix-noisy channel hybrid system. *Egypt Comput Sci J*, 40(2):54–64, 2016.

[32] Majed M Al-Jefri and Sabri A Mahmoud. Context-sensitive Arabic spell checker using context words and n-gram language models. In *2013 Taibah University International Conference on Advances in Information Technology for the Holy Quran and Its Sciences*, pages 258–263. IEEE, 2013.

[33] Chiraz Ben Othmane Zribi and Mohamed Ben Ahmed. Detection of semantic errors in Arabic texts. *Artificial Intelligence*, 195:249–264, 2013.

[34] Mahmoud Rokaya. Arabic semantic spell checking based on power links. *International Information Institute (Tokyo). Information*, 18(11):4749, 2015.

[35] Daniel Watson, Nasser Zalmout, and Nizar Habash. Utilizing character and word embeddings for text normalization with sequence-to-sequence models. *arXiv preprint arXiv:1809.01534*, 2018.

[36] Manar Alkhatib, Azza Abdel Monem, and Khaled Shaalan. Deep learning for Arabic error detection and correction. *ACM Transactions on Asian and Low-Resource Language Information Processing (TALLIP)*, 19(5):1–13, 2020.

[37] Aiman Solyman, Wang Zhenyu, Tao Qian, Arafat Abdulgader Mohammed Elhag, Muhammad Toseef, and Zeinab Aleibeid. Synthetic data with neural machine translation for automatic correction in arabic grammar. *Egyptian Informatics Journal*, 2020.

[38] Alex Kuznetsov and Hector Urdiales. Spelling correction with denoising transformer. *arXiv preprint arXiv:2105.05977*, 2021.

[39] Sepp Hochreiter and Jürgen Schmidhuber. Long short-term memory. *Neural Computation*, 9(8):1735–1780, 1997.

[40] Wikimedia_Commons. Long short-term memory, 2007.

[41] Gheith A. Abandah, Mohammed Z. Khedher, Mohammad R. Abdel-Majeed, Hamdi M Mansour, Salma F Hulliel, and Lara M Bisharat. Classifying and diacritizing Arabic poems using deep recurrent neural networks. *Journal of King Saud University - Computer and Information Sciences*, 2020.

[42] Ikbel Hadj Ali, Zied Mnasri, and Zied Lachiri. Dnn-based grapheme-to-phoneme conversion for arabic text-to-speech synthesis. *International Journal of Speech Technology*, 23(3):569–584, 2020.

[43] Nizar Habash, Anas Shahrour, and Muhamed Al-Khalil. Exploiting arabic diacritization for high quality automatic annotation. In *Proceedings of the Tenth International Conference on Language Resources and Evaluation (LREC'16)*, pages 4298–4304, 2016.

[44] Ibrahim Bounhas, Nadia Soudani, and Yahya Slimani. Building a morpho-semantic knowledge graph for arabic information retrieval. *Information Processing & Management*, 57(6):102124, 2020.

[45] Abdulrahman Alosaimy and Eric Atwell. Diacritization of a highly cited text: a classical arabic book as a case. In *2018 IEEE 2nd International Workshop on Arabic and Derived Script Analysis and Recognition (ASAR)*, pages 72–77. IEEE, 2018.

[46] Adel Abdelli, Fayçal Guerrouf, Okba Tibermacine, and Belkacem Abdelli. Sentiment analysis of arabic algerian dialect using a supervised method. In *2019 International Conference on Intelligent Systems and Advanced Computing Sciences (ISACS)*, pages 1–6. IEEE, 2019.

[47] Rajae Moumen, Raddouane Chiheb, Rdouan Faizi, and Abdellatif El Afia. Arabic diacritization with gated recurrent unit. In *Proceedings of the International Conference on Learning and Optimization Algorithms: Theory and Applications*, pages 1–4, 2018.

[48] Mokthar Ali Hasan Madhfar and Ali Mustafa Qamar. Effective deep learning models for automatic diacritization of arabic text. *IEEE Access*, 9:273–288, 2020.

[49] Ali Fadel, Ibraheem Tuffaha, Mahmoud Al-Ayyoub, et al. Arabic text diacritization using deep neural networks. In *2019 2nd International Conference on Computer Applications & Information Security (ICCAIS)*, pages 1–7. IEEE, 2019.

[50] Mohamed Maamouri, Ann Bies, Tim Buckwalter, and Wigdan Mekki. The Penn Arabic treebank: Building a large-scale annotated Arabic corpus. In *NEMLAR Conference on Arabic Language Resources and Tools*, volume 27, pages 466–467. Cairo, 2004.

[51] Imed Zitouni, Jeffrey Sorensen, and Ruhi Sarikaya. Maximum entropy based restoration of Arabic diacritics. In *Proceedings of the 21st International Conference on Computational Linguistics and 44th Annual Meeting of the Association for Computational Linguistics*, pages 577–584, 2006.

[52] Maha M Alamri and William J Teahan. Automatic correction of Arabic dyslexic text. *Computers*, 8(1):19, 2019.

[53] Yo Joong Choe, Jiyeon Ham, Kyubyong Park, and Yeoil Yoon. A neural grammatical error correction system built on better pre-training and sequential transfer learning. *arXiv preprint arXiv:1907.01256*, 2019.